\documentclass[10pt,twocolumn,letterpaper]{article}
\usepackage{authblk}
\usepackage{balance}

\usepackage[]{iccv}      % To produce the REVIEW version
\usepackage{graphicx}
\usepackage{amsmath}
\usepackage{amssymb}
\usepackage{booktabs}
\usepackage{graphics}
\usepackage{times}
\usepackage{upgreek}
\usepackage[dvipsnames,table,x11names]{xcolor}
\usepackage{tabularx,booktabs}
\usepackage{arydshln}
\usepackage{tabu}
\usepackage{adjustbox}
\usepackage{enumerate}
\usepackage{multirow}
\usepackage{multicol}
\usepackage{enumitem}
\usepackage{subcaption}

\usepackage{caption}
\usepackage{subcaption}
\usepackage{balance}
\usepackage{xspace}

\usepackage{pifont}
\usepackage{subcaption}
\usepackage{caption}
\usepackage{subcaption}
\usepackage{mathtools}
\usepackage{xfrac}
\usepackage{soul}
\usepackage{url}
\usepackage[linesnumbered,ruled]{algorithm2e}

\usepackage{algorithm}
\usepackage{algpseudocode}
\usepackage{cancel}
\definecolor{shadecolor}{rgb}{0.97,0.97,0.97}
\DeclareMathOperator*{\argmax}{argmax}
\usepackage{bbding}

\definecolor{iccvblue}{rgb}{0.21,0.49,0.74}
\usepackage[pagebackref,breaklinks,colorlinks,allcolors=iccvblue]{hyperref}

\usepackage[capitalize]{cleveref}
\crefname{section}{Sec.}{Secs.}
\Crefname{section}{Section}{Sections}
\Crefname{table}{Table}{Tables}
\crefname{table}{Tab.}{Tabs.}

\newif\ifdraft
\drafttrue

\definecolor{orange}{rgb}{1,0.5,0}
\definecolor{pink}{rgb}{0.98, 0.38, 0.5}
\definecolor{darkgreen}{rgb}{0.055, 0.490, 0.016} 

\ifdraft
 \usepackage[normalem]{ulem}
 \newcommand{\RS}[1]{{\color{red}{\bf RS: #1}}}
 
 \newcommand{\PMN}[1]{{\color{orange}{\bf PMN: #1}}}
 
 \newcommand{\JGT}[1]{{\color{blue} JGT: #1}}

\else
 \newcommand{\sout}[1]{}
 \newcommand{\RS}[1]{{\color{red}{}}}
 
 \newcommand{\PMN}[1]{{\color{red}{}}}
 
 \newcommand{\JGT}[1]{{\color{blue}{}}}
 
\fi

\newcolumntype{S}{>{\centering\arraybackslash}p{1.05cm}}
\newcolumntype{L}{>{\small}l}
\newcolumntype{T}{>{\centering\arraybackslash}p{1.1cm}}

\newcommand{\comment}[1]{}

\newcommand{\grayref}[1]{\textcolor{gray}{#1}}

\title{Feedback-Driven Pseudo-Label Reliability Assessment: Redefining Thresholding for Semi-Supervised Semantic Segmentation}

\author[1]{Negin Ghamsarian}
\author[2]{Sahar Nasirihaghighi}
\author[2]{Klaus Schoeffmann}
\author[1]{Raphael Sznitman}

\affil[1]{University of Bern, Bern, Switzerland}
\affil[2]{University of Klagenfurt, Klagenfurt, Austria}

\affil[ ]{\tt\small\{negin.ghamsarian, raphael.sznitman\}@unibe.ch, \{sahar.nasirihaghighi, klaus.schoeffmann\}@aau.at}
\begin{document}
\maketitle
\enlargethispage{\baselineskip}
\begin{abstract}
Semi-supervised learning leverages unlabeled data to enhance model performance, addressing the limitations of fully supervised approaches. Among its strategies, pseudo-supervision has proven highly effective, typically relying on one or multiple teacher networks to refine pseudo-labels before training a student network. A common practice in pseudo-supervision is filtering pseudo-labels based on pre-defined confidence thresholds or entropy. However, selecting optimal thresholds requires large labeled datasets, which are often scarce in real-world semi-supervised scenarios. To overcome this challenge, we propose Ensemble-of-Confidence Reinforcement (ENCORE), a dynamic feedback-driven thresholding strategy for pseudo-label selection. Instead of relying on static confidence thresholds, ENCORE estimates class-wise true-positive confidence within the unlabeled dataset and continuously adjusts thresholds based on the model’s response to different levels of pseudo-label filtering. This feedback-driven mechanism ensures the retention of informative pseudo-labels while filtering unreliable ones, enhancing model training without manual threshold tuning. Our method seamlessly integrates into existing pseudo-supervision frameworks and significantly improves segmentation performance, particularly in data-scarce conditions. Extensive experiments demonstrate that integrating ENCORE with existing pseudo-supervision frameworks enhances performance across multiple datasets and network architectures, validating its effectiveness in semi-supervised learning.
\end{abstract}
\section{Introduction}
\label{sec:introduction}

Supervised deep learning has demonstrated remarkable success in computer vision tasks, often surpassing classical machine learning methods. However, this success is contingent on large-scale annotated datasets, which are costly and time-intensive to obtain. Semantic segmentation, in particular, demands extensive pixel-wise annotations, making it significantly more expensive than region- or image-level annotation~\cite{MiCOCO}. Even state-of-the-art segmentation networks~\cite{DeepLab,DeepLabV3,PSPNet,unetpp,FCN,Mask-R-CNN} suffer from performance degradation in low-data regimes. The challenge is even more pronounced in medical image segmentation, where annotations require domain expertise and are highly labor-intensive.

To mitigate this reliance on labeled data, semi-supervised learning (SSL) has gained traction by leveraging large amounts of unlabeled data to improve generalization and domain adaptability~\cite{C3semiseg,DLDM,kalluri2019universal,BLDA,SSGM}. Among SSL techniques, pseudo-labeling is a widely adopted strategy in semantic segmentation~\cite{Pseudo-label}, where a model trained on labeled data generates approximate ground-truth labels for unlabeled images, incorporating them into training.

Pseudo-supervision methods typically employ a pretrained network \cite{STPP}, a mean teacher \cite{MT}, or multiple teacher models \cite{DualTeacher,AD-MT,GPS} to enhance pseudo-labeling. To mitigate error propagation, pseudo-labels undergo a filtering stage to exclude low-confidence predictions. However, existing confidence-based filtering approaches face limitations: high-confidence thresholds (e.g., 0.95) risk discarding numerous correct pseudo-labels, while hyperparameter search for threshold tuning contradicts the real-world constraints of SSL, where labeled data is scarce and a sufficiently large validation set is impractical.  

\begin{figure}[!t]
    \centering    \includegraphics[width=0.49\textwidth]{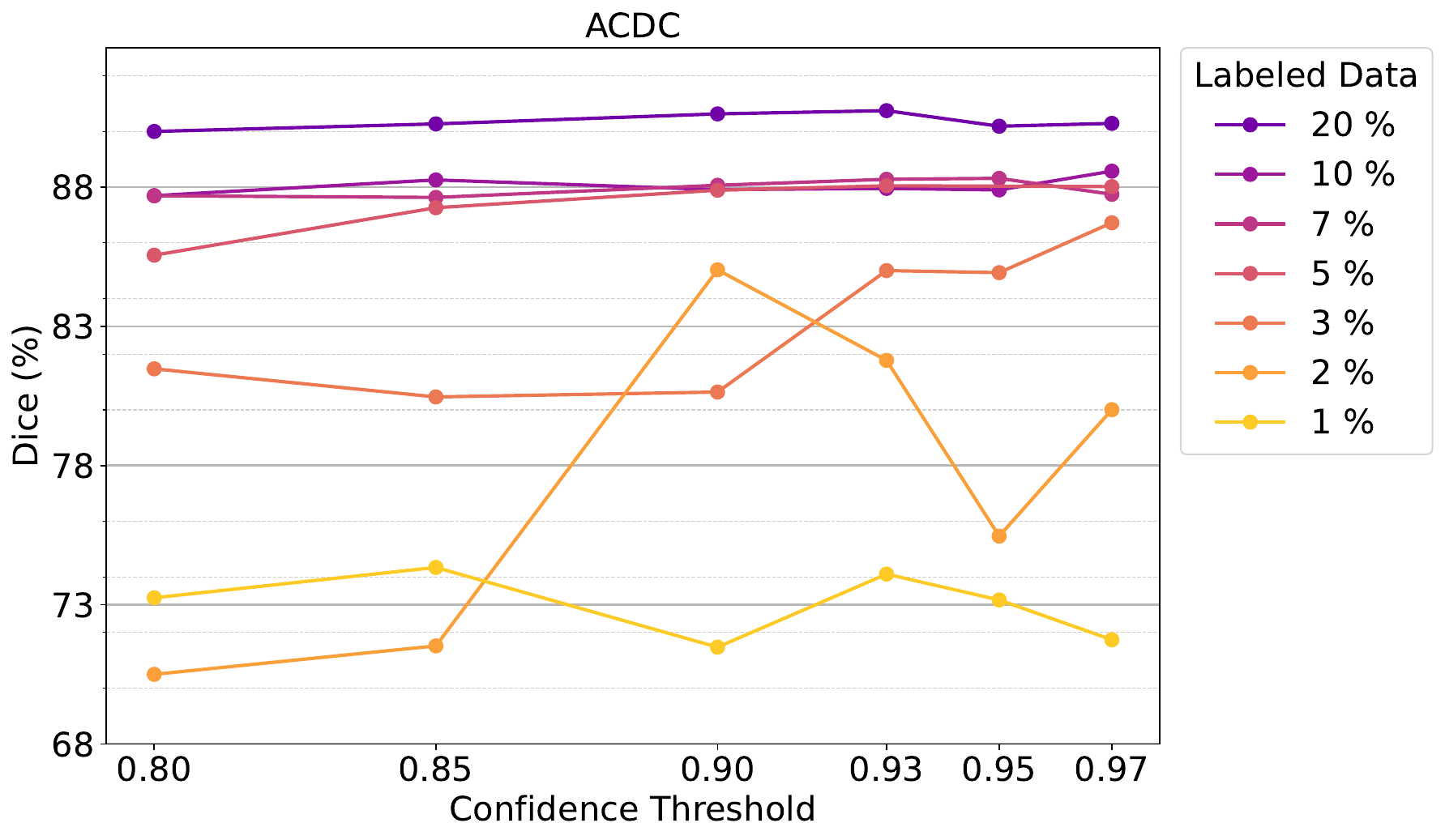}    \includegraphics[width=0.5\textwidth]{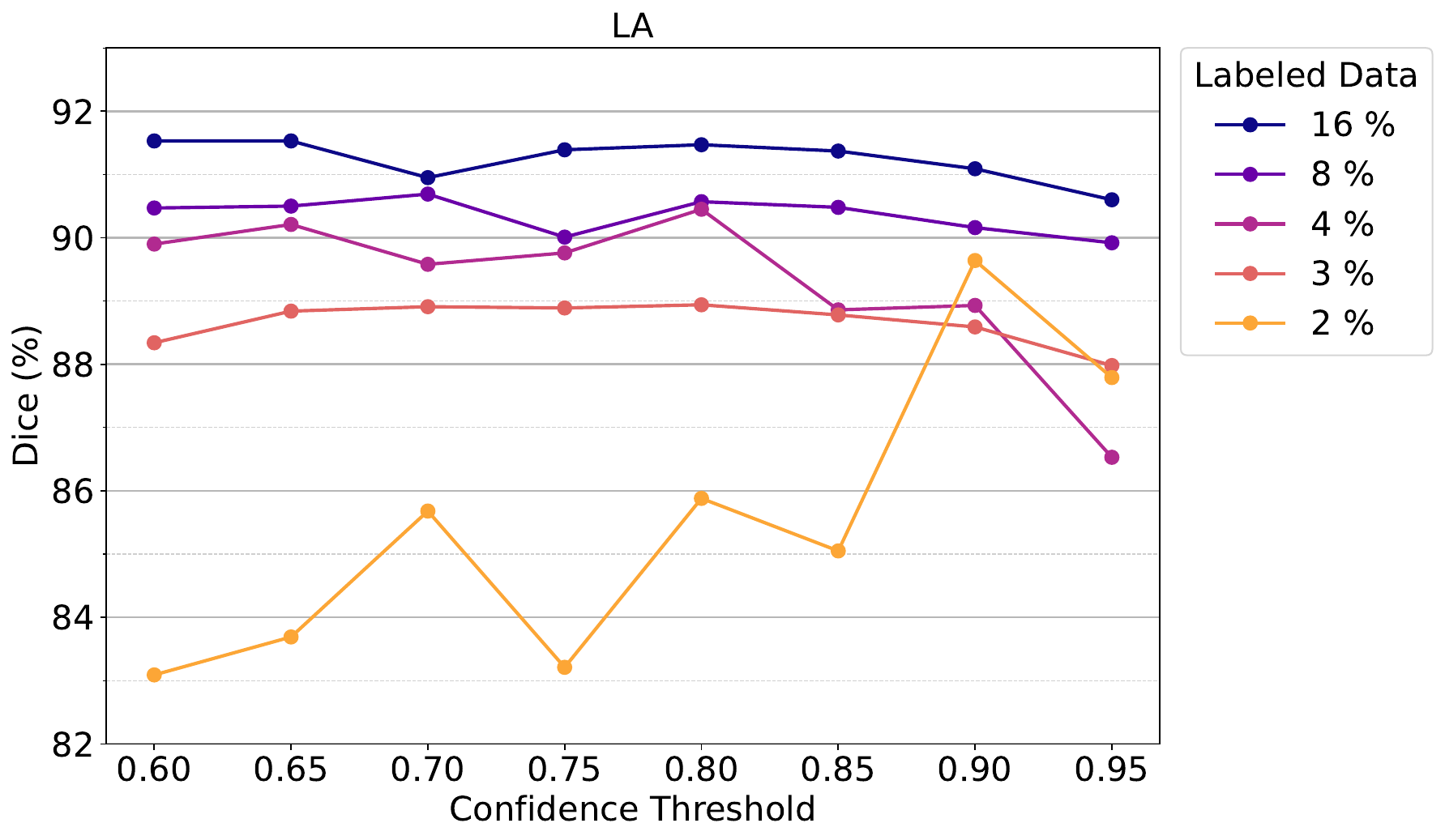}
    \caption{Comparison of Dice scores across different confidence thresholds for various labeled data percentages on the ACDC and LA datasets with AD-MT \cite{AD-MT}. The plots illustrate how segmentation performance fluctuates with threshold selection, particularly in low-data regimes. As the labeled data fraction decreases, model sensitivity to threshold choice increases, making optimal threshold selection more challenging. Notably, a fixed global threshold often fails to generalize well across different dataset conditions, underscoring the need for an adaptive thresholding strategy.}

    \label{fig:ThresholdVariation}
\end{figure}

Moreover, in low-data regimes, model performance is highly sensitive to the choice of confidence threshold, yet this dependency follows no consistent pattern (see Figure \ref{fig:ThresholdVariation}). As labeled dataset size decreases, the optimal threshold becomes increasingly unpredictable, significantly affecting segmentation performance. A fixed threshold may be too strict, discarding useful pseudo-labels, or too lenient, introducing noisy predictions. This issue is exacerbated by inter-class variability in confidence levels, where different classes exhibit varying prediction certainty, further complicating threshold optimization.

To address these challenges, we propose \textit{Ensemble-of-Confidence Reinforcement} (ENCORE), a two-stage pseudo-label assessment strategy that integrates \textit{Class-Aware Confidence Calibration (CAC)} and \textit{Adaptive Confidence Thresholding (ACT)}. CAC estimates class-wise pseudo-label confidence using only the available labeled dataset, ensuring that pseudo-label filtering accounts for inter-class confidence variability. ACT then dynamically adjusts confidence thresholds throughout training based on real-time feedback, eliminating the need for manual threshold tuning. Specifically, our contributions are as follows:

\begin{itemize}
    \item We introduce Class-Aware Confidence Calibration (CAC), which computes class-wise confidence thresholds by estimating the model's average true-positive confidence per class. This accounts for class-dependent variability in prediction confidence, ensuring more reliable pseudo-label selection. Accordingly, we utilize CAC to initialize class-wise pseudo-label filtering thresholds without requiring additional validation data or hyperparameter tuning.
    \item We propose Adaptive Confidence Thresholding (ACT), a dynamic threshold adjustment mechanism that continuously refines pseudo-label filtering during training. ACT evaluates the student model’s response to different thresholds using labeled batches and iteratively selects the most effective threshold for each stage of training.
    \item Our fully automated thresholding strategy, implemented within ENCORE, eliminates a major hyperparameter in pseudo-supervision while significantly enhancing semi-supervised learning performance across different pseudo-labeling frameworks.
\end{itemize}

The remainder of this paper is structured as follows: Section~\ref{sec:related work} reviews recent advancements in semi-supervised semantic segmentation. Section~\ref{sec:method} details our proposed framework, \textit{Ensemble of Confidence Reinforcement} (ENCORE). The experimental setup is described in Section~\ref{sec:Experimental Setup}, followed by experimental results and analysis in Section~\ref{sec:experimental results}. Finally, Section~\ref{sec:conclusion} summarizes our findings and discusses future research directions.

\begin{figure*}[t!]
    \centering    \includegraphics[width=0.83\textwidth]{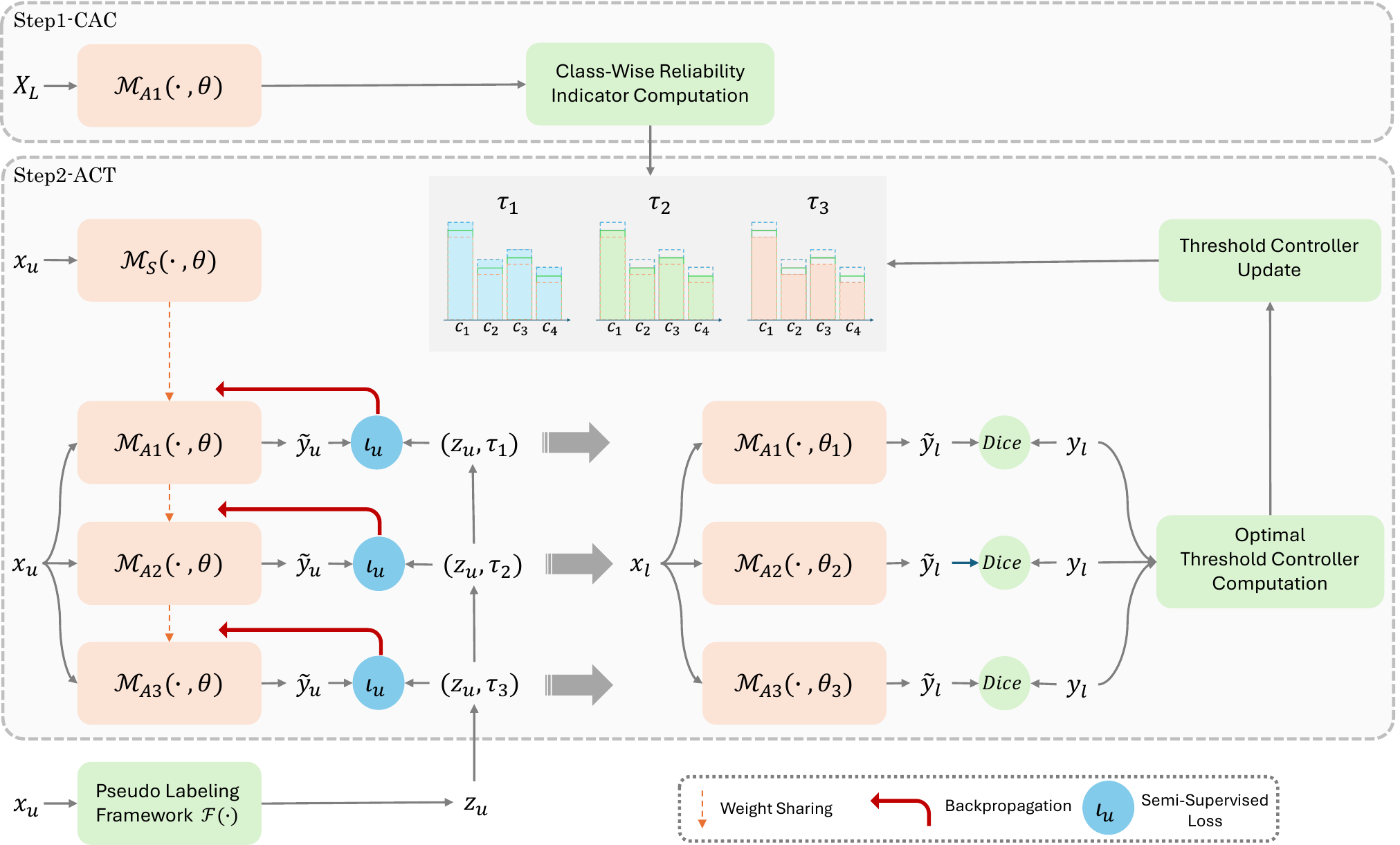}
    \caption{Overview of ENCORE: A feedback-driven pseudo-label refinement framework integrating Class-Aware Confidence Calibration (CAC) and Adaptive Confidence Thresholding (ACT) to enhance pseudo-label reliability and improve semi-supervised segmentation.}
    \label{fig:Overview}
\end{figure*}
\section{Related work}
\label{sec:related work}

\noindent\textbf{Semi-Supervised Learning (SSL)} aims to enhance model performance while minimizing annotation efforts by leveraging unlabeled data during training. SSL approaches can be broadly categorized into three paradigms: (i) \textit{contrastive learning}, (ii) \textit{consistency regularization}~\cite{MT,TTUDA,TESSL,UDAMIS,Fixmatch,PSMT,GCT,DCAC,CCT}, and (iii) \textit{pseudo-supervision}~\cite{UDACB,BLDA,Pseudoseg,CPS,MBDA,DSP,DACS,Cutmix,ClassMix,CowMix,STPP,U2PL,copy-paste,Adaptive-copy-paste, DMT}.  

\textit{Contrastive Learning} focuses on robust representation learning. This technique enhances SSL by maximizing the similarity between related data instances while minimizing it for unrelated ones, improving feature consistency~\cite{Adaptive-copy-paste, PCC, COMP, ECIP}. By enforcing high agreement among similar samples and low agreement among dissimilar pairs, contrastive learning effectively captures the underlying structure of the input data, further improving segmentation performance in SSL frameworks. 

\textit{Consistency regularization} enforces stable model predictions under different conditions, ensuring robustness against variations such as (a) \textit{input perturbations}, including data transformations~\cite{RSTP,UDA,DCAC,DualTeacher,AD-MT,WS}, (b) \textit{network perturbations}, such as stochastic initializations~\cite{PSMT}, and (c) \textit{latent space augmentations}~\cite{PSMT} and drop-out \cite{WS}. Additionally, temporal consistency strategies enforce stability across different training checkpoints, reducing prediction variance~\cite{TESSL}.

\noindent\textbf{Pseudo-Supervision} enhances SSL by expanding the training set using model-generated pseudo-labels for unlabeled data. This method follows three primary paradigms: self-training, co-training, and mean teacher models.  
\textit{Self-training} relies on a model to generate pseudo-labels for unlabeled samples, retaining only high-confidence predictions for subsequent training~\cite{STPP,Pseudoseg,Crest,UDACB,copy-paste,Cutmix,Sun_2024_CVPR}. In contrast, \textit{co-training} employs multiple models trained on different data perspectives, where each model generates pseudo-labels to iteratively improve the other’s predictions~\cite{CPS,DMT}, enhancing robustness and generalization.  

Pseudo-supervision methods primarily focus on three challenges:  
1) Improving pseudo-label reliability,  
2) Mitigating confirmation bias, which arises when incorrect pseudo-labels reinforce model errors, and  
3) Adapting pseudo-label incorporation strategies to improve model generalization, particularly in low-data regimes.

\section{Method}
\label{sec:method}
In this section, we formally define the problem and introduce our proposed thresholding framework for semi-supervised semantic segmentation. We begin with a mathematical formulation of the problem, followed by an overview of the proposed approach. Subsequently, we provide a detailed explanation of its two key components in Sec.~\ref{subsec: CAC} and Sec.~\ref{subsec: ACT}.

\noindent \textit{\textbf{Problem Definition. }}
\label{subsec: problem definition}
We consider a labeled dataset \( \mathcal{D^L} = \{(x_l^{i}, y_l^{i})\}_{i=1}^{N^\mathcal{L}} \), consisting of training images \( \mathcal{X} \) and their corresponding segmentation labels \( \mathcal{Y} \), as well as an unlabeled dataset \( \mathcal{D^U} = \{x_u^{i}\}_{i=1}^{N^\mathcal{U}} \), containing only the unlabeled image set \( \mathcal{U} \), where \( N^\mathcal{U} \gg N^\mathcal{L} \). Our objective is to train a model \( \mathcal{M}_S(\cdot, \theta) \) that leverages both labeled and unlabeled data to improve segmentation performance on unseen samples. To achieve this, we adopt a semi-supervised learning paradigm, where pseudo-labels are assigned to the images in \( \mathcal{U} \) using a pseudo-supervision framework. The overall training objective is defined as \( \mathcal{L}_{\text{overall}} =  \mathcal{L}_{l} + \lambda \mathcal{L}_{u} \), where \( \mathcal{L}_{l} \) is the supervised loss on labeled data, \( \mathcal{L}_{u} \) is the unsupervised loss on pseudo-labeled data, and \( \lambda \) is a framework-dependent weighting hyperparameter that balances the contribution of the two terms. The core challenge is to identify and filter out unreliable pseudo-labels when computing \( \mathcal{L}_{u} \) to maximize segmentation performance.

\noindent \textit{\textbf{Overview. }} Figure~\ref{fig:Overview} illustrates the framework of our proposed semi-supervised semantic segmentation method, termed ``Ensemble-of-Confidence Reinforcement" (ENCORE). The method introduces a feedback-driven, adaptive thresholding framework that improves pseudo-label reliability. The student network (\(\mathcal{M}_S(\cdot, \theta)\)) learns from both labeled and pseudo-labeled data, while a set of assessor networks (\(\{\mathcal{M}_{Ai}(\cdot, \theta_i)\}_{i=1}^3\)) dynamically evaluate pseudo-label reliability based on feedback derived from the student’s response to labeled images. At the core of our approach are two key innovations: \textit{Class-Aware Confidence Calibration (CAC)} and \textit{Adaptive Confidence Thresholding (ACT)}. CAC refines pseudo-label selection by computing class-wise reliability scores, ensuring that confidence thresholds are adjusted per class rather than relying on a static global threshold. ACT further improves pseudo-label selection by iteratively evaluating different threshold levels using assessor networks and selecting the optimal threshold based on segmentation performance on labeled data. Unlike conventional methods with fixed confidence thresholds, ENCORE continuously adapts its thresholding strategy based on real-time feedback, preventing over- or under-filtering of pseudo-labels and enhancing segmentation accuracy. By integrating CAC and ACT, our approach dynamically calibrates confidence levels and optimally selects pseudo-labels, leading to more reliable supervision and improved segmentation performance.

\begin{figure}[!b]
    \centering    \includegraphics[width=0.45\textwidth]{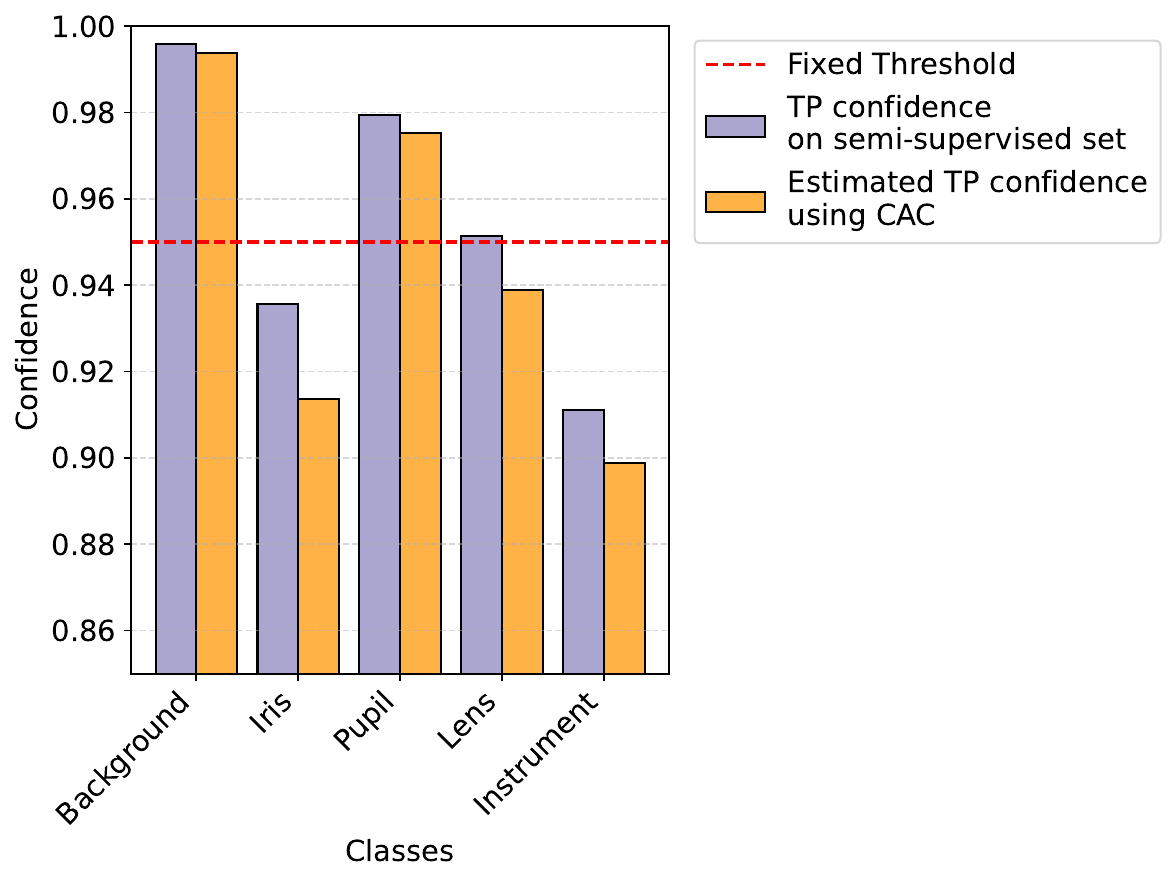}    
    \caption{Comparison between CAC-based confidence thresholds and fixed thresholds for Cataract-1K dataset (1/8 labeled data).}

    \label{fig:CAC}
\end{figure}

\subsection{Class-Aware Confidence Calibration (CAC)}
\label{subsec: CAC}

Standard pseudo-label confidence indicators, such as the maximal softmax probability, often suffer from calibration issues, leading to suboptimal pseudo-label selection. Conventional methods like UniMatch~\cite{DualTeacher,AD-MT,GPS,WS,MT,MCF,PSMT} apply a single, global confidence threshold \( \tau \) uniformly across all classes. However, this one-size-fits-all approach fails to account for class-specific score distributions. As a result, setting \( \tau \) too high disproportionately filters out correct pseudo-labels for inherently low-confidence classes, increasing false negatives. Conversely, lowering \( \tau \) to accommodate these classes introduces excessive false positives for high-confidence classes. This highlights a critical limitation: a fixed confidence threshold fails to capture class-wise prediction reliability, motivating the need for a more adaptive, class-sensitive confidence measure.  

To address this issue, we propose Class-Aware Confidence Calibration (CAC), which estimates prediction reliability on a per-class basis. CAC utilizes the labeled training set as a reference to assess how model confidence correlates with ground-truth correctness for each class. Instead of applying a uniform threshold, CAC dynamically determines class-specific confidence levels by computing the \textit{average confidence of true positive detections} (Figure \ref{fig:CAC}).  

Formally, let \( \mathcal{Y} = \{y_l^i\}_{i=1}^{N^\mathcal{L}} \) denote the ground-truth labels, and let \( \mathcal{M}_A(\mathcal{X}) = \{\mathcal{M}_A(x_l^i)\}_{i=1}^{N^\mathcal{L}} \) represent the network outputs for the labeled images. The predicted label for each image is then given by:

\begin{equation}
    \hat{y}_l^i = \arg\max_C\Big(\tilde{y}_l^i\Big),
\end{equation}

where \( \tilde{y}_l^i \in \mathbb{R}^{C \times W \times H} \) represents the softmax probabilities across all classes, defined as \( \tilde{y}_l^i = \sigma\big(\mathcal{M_A}(x_l^i)\big) \), with \( \sigma \) being the softmax operator.  
The \textit{class-wise reliability indicator} \( R \) is then computed as:

\begin{equation}
\begin{aligned}
    R &= [R_1, R_2, ..., R_C], \\
    R_{c \in \{1,C\}} &= \frac{1}{N^\mathcal{L}} \sum_{i=1}^{N^\mathcal{L}} \frac{1}{N_i^c} \sum_{j=1}^k \Big([\hat{y}_l^{ij} = c] \times [y_l^{ij} = c]\Big), \\
    N_i^c &= \sum_{j=1}^k [y_l^{ij} = c],
\end{aligned}
\label{eq:reliability_indicator}
\end{equation}

where \( C \) is the number of classes, \( k \) represents the number of pixels in the segmentation mask \( y_l \), and \( [P] \) denotes the Iverson bracket, which evaluates to 1 if \( P \) is true and 0 otherwise.  

The \textit{class-wise reliability indicator \( R \) is computed once} using a network trained solely on labeled images and is set as the initial class-wise confidence threshold in the first iteration of semi-supervised learning. This enables a more adaptive pseudo-label selection strategy, ensuring that confidence thresholds are class-specific rather than globally fixed, thereby reducing bias toward high-confidence classes and improving segmentation performance across all classes.

\subsection{Adaptive Confidence Thresholding (ACT)}
\label{subsec: ACT}

While the CAC module mitigates inter-class confidence variability, it does not inherently ensure that filtering pseudo-labels based on the reliability indicator maximizes segmentation performance. The optimal balance between bridging the distribution gap and minimizing false positives depends on various factors, including dataset characteristics, the fraction of labeled data, the network’s training progression (e.g., epoch number), and the underlying semi-supervised framework. Instead of relying on a fixed class-wise confidence threshold, we propose an adaptive threshold selection strategy that evaluates multiple threshold levels, compares their performance, and dynamically selects the most suitable one.
To achieve this, we initialize three class-wise confidence threshold controllers as follows:

\begin{equation}
    (\tau_1, \tau_2, \tau_3) = (\alpha_1 R, R, \alpha_2 R),
    \label{eq:init controllers}
\end{equation}

where \( R \) represents the \textit{class-wise reliability indicator} computed via CAC, and \( \alpha_1, \alpha_2 \) are \textit{threshold adaptors} that define the initial search space.

At each iteration, three assessor networks are instantiated by copying the current student model weights. Each assessor network applies a different threshold controller to filter pseudo-labels before training on the pseudo-labeled batch. The segmentation performance of each assessor is then evaluated on the labeled batch using the Dice metric, and the threshold controller that yields the highest Dice score is selected as the optimal controller for pseudo-label filtering in the next student model training cycle:

\begin{equation}
    \tau^* = \argmax_{\tau_k \in \{\tau_1, \tau_2, \tau_3\}} \text{Dice}(\mathcal{M}_{A_k}(x_l; \theta_k), y_l, \tau_k),
    \label{eq:optimal controller}
\end{equation}

where \( \mathcal{M}_{A_k} \) denotes the assessor network trained with threshold \( \tau_k \), and \( \text{Dice}(\cdot, \cdot, \tau_k) \) evaluates segmentation performance with threshold \( \tau_k \).

As training progresses, the recurrent selection of a particular threshold controller provides insight into whether the confidence thresholds should be dynamically adjusted. If the lowest or highest confidence threshold is consistently chosen, this suggests that the model's overall confidence levels should be shifted downward or upward, respectively. The threshold controllers are updated according to:

\begin{equation}
    (\tau_1, \tau_2, \tau_3) =
    \begin{cases} 
        (\alpha_1 \tau_1, \tau_1, \alpha_2 \tau_1), & \text{if } \tau^* = \tau_1 \text{ and } N_{\tau^*} \geq n, \\
        (\alpha_1 \tau_3, \tau_3, \alpha_2 \tau_3), & \text{if } \tau^* = \tau_3 \text{ and } N_{\tau^*} \geq n, \\
        (\tau_1, \tau_2, \tau_3), & \text{else}.
    \end{cases}
    \label{eq:update controller}
\end{equation}

Here, \( \alpha_1 \) and \( \alpha_2 \) determine how aggressively the confidence thresholds are adjusted. To prevent excessive adaptation, we introduce a mechanism that modifies the threshold controllers only if the same threshold controller is selected more than \( n \) consecutive times.
Hence, \( N_{\tau^*} \) tracks the number of consecutive selections of \( \tau^* \).  

Although \( \alpha_1 \) and \( \alpha_2 \) are fixed hyperparameters, our dynamic feedback-driven adaptation eliminates manual tuning by iteratively refining confidence thresholds based on model performance. This metric-adaptive calibration ensures that pseudo-labels meet robustness criteria—such as the Dice metric—enhancing segmentation accuracy based on observed performance rather than predefined heuristics.

This approach adaptively refines threshold controllers, allowing the model to dynamically adjust confidence filtering in response to its learning progression, improving pseudo-label reliability throughout training.

\begin{algorithm}[t!]

    \SetKwInOut{Input}{Input}
    \SetKwInOut{Output}{Output}
    
    \Input{Labeled training set $\mathcal{D^L} = \{(x_l^i, y_l^i)\}_{i=1}^{N^\mathcal{L}}$,\\
    Unlabeled training set $\mathcal{D^U} = \{(x_u^i)\}_{i=1}^{N^\mathcal{U}}$,\\
    Student model $\mathcal{M}_S(\cdot, \theta)$,\\
    Pseudo-labeling framework $\mathcal{F}(\cdot)$,\\
    Assessor models $\mathcal{M}_{A1}$, $\mathcal{M}_{A2}$, $\mathcal{M}_{A3}$,
    }
    \Output{Trained student model $\mathcal{M}_S(\cdot, \theta)$.}
    
    Train an assessor network on labeled images. 
    
    Compute the class-wise reliability indicator. \Comment{Eq.~\eqref{eq:reliability_indicator}}

    Initialize confidence threshold controllers.
    \Comment{Eq.~\eqref{eq:init controllers}}
    
    \For{each mini-batch pair in ($\mathcal{D^L}$, $\mathcal{D^U}$)}{
        Compute pseudo labels for the unlabeled training batch $\Tilde{y}_u$=$\mathcal{F}(x_u)$,
        
        Load the assessor models with the current student weights.
        
        Train the assessor models on $x_u$ using the threshold controllers.
        
        Compute the optimal threshold controller. \Comment{Eq.~\eqref{eq:optimal controller}}
        
        Train the student model with optimal threshold controller.
        
        Update threshold controllers.
        \Comment{Eq.~\eqref{eq:update controller}}
        }

    \caption{Ensemble-of-Confidence Reinforcement for Semi-Supervised Semantic Segmentation}
    \label{algorithm}
\end{algorithm}

\section{Experimental setup}
\label{sec:Experimental Setup}

\paragraph{Dataset. }We evaluate the proposed and baseline methods on five widely used benchmarks for medical image segmentation from different modalities: 
\begin{itemize}
    \item \textbf{Cataract-1K} dataset \cite{Cataract-1K}, which consists of the annotations of the eye's relevant anatomical objects including iris, pupil, intraocular lens, plus instruments in 2256 frames out of 30 cataract surgery videos. For semi-supervised settings, we randomly select six cases as the test set, $[13, 6, 3, 1]$ cases (corresponding to $[1/2, 1/4, 1/8, 1/16]$ splits) out of the training set as the supervised set, and the remaining cases will be used as the unlabeled semi-supervised set.

    \item \textbf{Prostate MRI} dataset \cite{s-net} includes 116 prostate MRI volumes from six different sites. For each fold, 16 volumes are randomly sampled as the test set, and $[1/2, 1/4, 1/8, 1/14, 1/25, 1/50]$ of the cases are randomly sampled as the labeled set.

    \item \textbf{EndoVis} dataset \cite{EndoVis} includes annotations of instruments from 73 endoscopic surgery videos. We split these videos in a patient-wise manner to the training and testing set. Afterward, the training set is split into supervised and semi-supervised sets, with the rate of supervised images being equal to $[1/2, 1/4, 1/8, 1/16, 1/32, 1/64, 1/128]$. 

    \item \textbf{ACDC} dataset \cite{bernard2018deep}, which includes 100 cine-MRI scans consisting of four classes (background, right ventricle, left ventricle, and myocardium). For this dataset, we adhere to all the settings outlined in \cite{AD-MT}, while excluding it from our configurations. The training set is splited into different volumes at ratios of $[1/7, 1/14, 1/70, 1/140]$, corresponding to 262, 136, 26, and 13 slices, respectively. 
    
    \item \textbf{LA} dataset \cite{xiong2021global}, which comprises 100 3D late gadolinium-enhanced (LGE) MR scans with corresponding left atrium segmentation mask. In accordance with the task settings described in \cite{AD-MT}, we partitioned the dataset into 80 training scans—with volume splits of $[1/10, 1/20, 1/26, 1/40]$— and 20 testing scans, applying identical preprocessing procedures across the board.
\end{itemize}

\vspace{-0.8\baselineskip}
\paragraph{Implementation Details. }
For LCDC and LA datasets, we follow all settings of AD-MT~\cite{AD-MT}. Specifically, we use UNet for ACDC and VNet for LA dataset. For remaining datasets, we utilize DeepLabV3+~\cite{DeepLabV3} with a ResNet50 backbones~\cite{ResNet} initialized with the ImageNet~\cite{ImageNet} pre-trained parameters. Besides, the network includes dropout layers in the decoder. For fair comparisons, the hyperparameters are maintained the same for all semi-supervised methods. We use stochastic gradient descent for optimization.  The initial learning rate ($lr_{init}$) of the randomly-initialized segmentation head is set to 0.005 for Cataract-1K, and 0.001 for other datasets. The learning rate of the pre-trained encoder is $10\%$ of that of the segmentation head. The learning rate is decreased during training using polynomial decay with the power of 0.9: $lr = lr_{init}\times (1-\frac{iter}{total-iter})$. All models are trained for 80 epochs on the Cataract-1K and 40 epochs on Prostate MRI and EndoVis datasets. For the loss function, we adhere to the loss formulation specified by each semi-supervised framework. As weak augmentations, all images are randomly rescaled, horizontally flipped with the probability of $0.5$, and cropped with a size of $(384, 384)$ for Cataract-1K, $(352, 256)$ for EndoVis, and $(256, 256)$ for other datasets. The strong augmentations include color-jittering (with an intensity factor of 0.6 for brightness and contrast, and 0.4 for saturation), and a random selection of Gaussian blur and random-adjust-sharpness. Besides, $(\alpha_1,\alpha_2)=(0.98, 1.02)$ and $n$ is set to five iterations across all settings.

% \vspace{-0.8\baselineskip}
% \paragraph{Augmentation. }

\vspace{-0.8\baselineskip}
\paragraph{Alternative methods. }We compare the performance of the proposed method with state-of-the-art semi-supervised learning frameworks including Mean Teacher \cite{MT}, CPS \cite{CPS}, RL \cite{RL}, ST++ \cite{STPP}, PS-MT \cite{PSMT}, MCF \cite{MCF}, BCP \cite{BCP}, UniMatch \cite{WS}, Switch \cite{DualTeacher}, GPS \cite{GPS}, and AD-MT \cite{AD-MT}. 

\vspace{-0.8\baselineskip}
\paragraph{Evaluation. } For all scenarios except ACDC and LA datasets, we employ a four-fold validation approach and present the averaged results across all folds. We evaluate the performance of all methods using the dice coefficient. All experiments are conducted on ``NVIDIA RTX: 3090" GPUs. To support reproducibility, the training splits for all datasets and codes will be released with the acceptance of the paper.
\section{Experimental Results}
\label{sec:experimental results}

\begin{table}[b!]
\centering
\caption{Comparisons with state-of-the-art methods on \textbf{Cataract-1K} test set (Network: DeepLabV3+) based on dice coefficient ($\%$).}
\label{tab:Cat1k}
\renewcommand{\arraystretch}{1.1}
\resizebox{1\columnwidth}{!}{%
\begin{tabular}{r*{1}@{\hspace{1.8pt}}l*{1}{>{\centering\arraybackslash}m{1.6cm}}*{4}{>{\centering\arraybackslash}m{1.6cm}}*{1}{>{\centering\arraybackslash}m{0.5cm}}}
\specialrule{.12em}{.05em}{.05em}
Framework &&1/2 (987)&1/4 (433)&1/8 (242)&1/26 (61)& Rel. Avg. \\
\specialrule{.12em}{.05em}{.05em} 
Supervised && $87.85 \; \grayref{\pm0.3}$& $82.55 \; \grayref{\pm1.9}$& $76.72 \; \grayref{\pm2.2}$& $60.34 \; \grayref{\pm3.9}$& N/A\\
\rowcolor{shadecolor}MT\cite{MT} & \grayref{[NeurIPS 2017]}& $87.10 \; \grayref{\pm1.0}$& $85.59 \; \grayref{\pm1.8}$& $84.57 \; \grayref{\pm0.7}$& $73.47 \; \grayref{\pm2.0}$& 5.82\\
CPS \cite{CPS} & \grayref{[CVPR 2021]}& $89.68 \; \grayref{\pm6.7}$& $84.77 \; \grayref{\pm6.6}$& $79.89 \; \grayref{\pm6.2}$& $64.63 \; \grayref{\pm8.2}$& 2.88\\
\rowcolor{shadecolor}RL \cite{RL} & \grayref{[MICCAI 2021]}& $85.74 \; \grayref{\pm1.2}$& $84.78 \; \grayref{\pm1.0}$& $83.33 \; \grayref{\pm0.5}$& $74.66 \; \grayref{\pm1.3}$& 5.26\\
ST++ \cite{STPP} & \grayref{[CVPR 2022]}& $88.73 \; \grayref{\pm0.2}$& $84.32 \; \grayref{\pm1.0}$& $80.74 \; \grayref{\pm0.9}$& $64.57 \; \grayref{\pm2.3}$& 2.73\\
\rowcolor{shadecolor}PS-MT \cite{PSMT} & \grayref{[CVPR 2022]}& $89.09 \; \grayref{\pm4.1}$& $86.44 \; \grayref{\pm4.7}$& $83.02 \; \grayref{\pm6.7}$& $62.94 \; \grayref{\pm7.3}$& 3.51\\
MCF \cite{MCF} & \grayref{[CVPR 2023]}& $89.21 \; \grayref{\pm7.3}$& $84.44 \; \grayref{\pm4.1}$& $79.57 \; \grayref{\pm1.6}$& $66.62 \; \grayref{\pm3.9}$& 3.10\\
\rowcolor{shadecolor}BCP \cite{BCP} & \grayref{[CVPR 2023]}& $87.23 \; \grayref{\pm0.3}$& $86.11 \; \grayref{\pm1.1}$& $81.86 \; \grayref{\pm1.9}$& $72.32 \; \grayref{\pm2.0}$& 5.01\\
GPS \cite{GPS} & \grayref{[WACV 2024]}& $89.48 \; \grayref{\pm0.5}$& $86.67 \; \grayref{\pm1.1}$& $81.24 \; \grayref{\pm1.4}$& $63.01 \; \grayref{\pm5.6}$& 3.24\\
\specialrule{.02em}{.05em}{.05em}
UniMatch \cite{WS} & \grayref{[CVPR 2023]}& $87.83 \; \grayref{\pm1.1}$& $82.98 \; \grayref{\pm3.8}$& $78.67 \; \grayref{\pm7.6}$& $78.07  \; \grayref{\pm8.4}$& 5.02\\
\rowcolor{shadecolor}UniMatch + ENCORE && $90.59 \; \grayref{\pm0.5}$& $90.20 \; \grayref{\pm0.2}$& $89.53 \; \grayref{\pm0.7}$& $83.62 \; \grayref{\pm2.4}$& 11.62\\
\specialrule{.02em}{.05em}{.05em}
Switch \cite{DualTeacher} & \grayref{[NeurIPS 2024]}& $90.15 \; \grayref{\pm0.7}$& $84.47 \; \grayref{\pm5.3}$& $81.24 \; \grayref{\pm4.6}$& $62.81 \; \grayref{\pm5.5}$& 2.80\\
\rowcolor{shadecolor}Switch + ENCORE && $90.78 \; \grayref{\pm0.4}$& $90.65 \; \grayref{\pm0.4}$& $89.39 \; \grayref{\pm0.4}$& $83.21 \; \grayref{\pm5.7}$& 11.64\\
\specialrule{.02em}{.05em}{.05em}
AD-MT \cite{AD-MT} & \grayref{[ECCV 2024]}& $89.03 \; \grayref{\pm2.7}$& $85.04 \; \grayref{\pm1.4}$& $78.46 \; \grayref{\pm1.6}$& $62.74 \; \grayref{\pm3.5}$& 1.95\\
\rowcolor{shadecolor}AD-MT + ENCORE && $89.39 \; \grayref{\pm0.4}$& $86.87 \; \grayref{\pm1.7}$& $84.33 \; \grayref{\pm1.4}$& $71.62 \; \grayref{\pm2.1}$& 6.19\\
\specialrule{.12em}{.05em}{.05em}
\end{tabular}
}
\end{table}

\begin{table}[bt!]
\centering
\caption{Comparisons with state-of-the-art methods on \textbf{Prostate MRI} test set (Network: DeepLabV3+) based on dice coefficient ($\%$).}
\label{tab:MRI}
\renewcommand{\arraystretch}{1.1}
\resizebox{\columnwidth}{!}{%
\begin{tabular}{r*{1}@{\hspace{2pt}}l*{1}{>{\centering\arraybackslash}m{1.6cm}}*{4}{>{\centering\arraybackslash}m{1.6cm}}*{1}{>{\centering\arraybackslash}m{1cm}}}
\specialrule{.12em}{.05em}{.05em}
Framework &&1/2 (1673)&1/4 (872)&1/8 (444)&1/14 (279)& Rel. Avg. \\
\specialrule{.12em}{.05em}{.05em} 
Supervised && $77.81 \; \grayref{\pm3.1}$& $74.46 \; \grayref{\pm3.1}$& $72.37 \; \grayref{\pm1.4}$& $65.05 \; \grayref{\pm4.9}$& N/A\\
\rowcolor{shadecolor}MT \cite{MT} & \grayref{[NeurIPS 2017]}& $82.47 \; \grayref{\pm1.3}$& $81.73 \; \grayref{\pm2.0}$& $79.71 \; \grayref{\pm1.7}$& $75.79 \; \grayref{\pm3.1}$& 8.42\\
CPS \cite{CPS} & \grayref{[CVPR 2021]}& $78.48 \; \grayref{\pm2.2}$& $76.94 \; \grayref{\pm4.4}$& $72.72 \; \grayref{\pm2.7}$& $64.78 \; \grayref{\pm4.5}$& 1.84\\
\rowcolor{shadecolor}RL \cite{RL} & \grayref{[MICCAI 2021]}& $79.87 \; \grayref{\pm1.7}$& $78.48 \; \grayref{\pm2.8}$& $77.97 \; \grayref{\pm2.7}$& $72.28 \; \grayref{\pm3.6}$& 7.12\\
ST++ \cite{STPP} & \grayref{[CVPR 2022]}& $79.44 \; \grayref{\pm1.3}$& $76.47 \; \grayref{\pm2.7}$& $75.36 \; \grayref{\pm1.7}$& $69.68 \; \grayref{\pm7.5}$& 2.54\\
\rowcolor{shadecolor}PS-MT \cite{PSMT} & \grayref{[CVPR 2022]}& $79.83 \; \grayref{\pm1.0}$& $77.41 \; \grayref{\pm3.8}$& $74.89 \; \grayref{\pm2.2}$& $69.42 \; \grayref{\pm4.6}$& 4.97\\
MCF \cite{MCF} & \grayref{[CVPR 2023]}& $76.96 \; \grayref{\pm1.6}$& $73.57 \; \grayref{\pm3.9}$& $73.51 \; \grayref{\pm1.8}$& $71.94 \; \grayref{\pm4.7}$& 4.07\\
\rowcolor{shadecolor}BCP \cite{BCP} & \grayref{[CVPR 2023]}& $80.16 \; \grayref{\pm1.4}$& $78.95 \; \grayref{\pm1.1}$& $74.65 \; \grayref{\pm2.4}$& $69.74 \; \grayref{\pm0.8}$& 3.87\\
GPS \cite{GPS} & \grayref{[WACV 2024]}& $83.94 \; \grayref{\pm0.9}$& $83.49  \; \grayref{\pm1.9}$& $81.17  \; \grayref{\pm1.9}$& $79.02  \; \grayref{\pm2.7}$& 10.73\\
\specialrule{.12em}{.05em}{.05em} 
UniMatch \cite{WS} & \grayref{[CVPR 2023]}& $79.96 \; \grayref{\pm1.0}$& $77.94 \; \grayref{\pm2.1}$& $74.42 \; \grayref{\pm1.9}$& $73.86 \; \grayref{\pm2.3}$& 4.82\\
\rowcolor{shadecolor}UniMatch + ENCORE &&  $82.45 \; \grayref{\pm0.8}$& $81.59 \; \grayref{\pm1.8}$& $78.85 \; \grayref{\pm4.5}$& $78.65 \; \grayref{\pm2.2}$& 9.91\\
\specialrule{.12em}{.05em}{.05em} 
Switch \cite{DualTeacher} & \grayref{[NeurIPS 2024]}& $79.45 \; \grayref{\pm0.4}$& $77.66 \; \grayref{\pm3.6}$& $75.72 \; \grayref{\pm2.8}$& $70.04 \; \grayref{\pm4.1}$& 4.79\\
Switch + ENCORE && $83.98 \; \grayref{\pm0.8}$& $84.24 \; \grayref{\pm1.6}$& $78.96 \; \grayref{\pm5.1}$& $77.89 \; \grayref{\pm4.7}$& 9.81\\
\specialrule{.12em}{.05em}{.05em} 
AD-MT \cite{AD-MT} & \grayref{[ECCV 2024]}& $82.21 \; \grayref{\pm1.3}$& $80.03 \; \grayref{\pm2.3}$& $78.26 \; \grayref{\pm2.0}$& $75.00 \; \grayref{\pm3.5}$& 7.67\\
\rowcolor{shadecolor}AD-MT + ENCORE &&  $83.08 \; \grayref{\pm0.9}$& $80.49 \; \grayref{\pm2.1}$& $80.21 \; \grayref{\pm1.0}$& $76.08 \; \grayref{\pm2.4}$& 8.42\\
\specialrule{.12em}{.05em}{.05em} 
\end{tabular}
}
\end{table}

\begin{table}[tb!]
\centering
\caption{Comparisons with state-of-the-art methods on \textbf{EndoVis} test set (Network: DeepLabV3+) based on dice coefficient ($\%$).}
\label{tab:EndoVis}
\renewcommand{\arraystretch}{1.1}
\resizebox{1\columnwidth}{!}{%
\begin{tabular}{r*{1}@{\hspace{2pt}}l*{1}{>{\centering\arraybackslash}m{1.6cm}}*{4}{>{\centering\arraybackslash}m{1.6cm}}*{1}{>{\centering\arraybackslash}m{1cm}}}
\specialrule{.12em}{.05em}{.05em}
Framework &&1/2 (1764)&1/4 (882)&1/8 (441)&1/16 (221)& Rel. Avg.\\
\specialrule{.12em}{.05em}{.05em} 
Supervised && $78.46 \; \grayref{\pm1.1}$& $75.40 \; \grayref{\pm1.2}$& $72.28 \; \grayref{\pm0.8}$& $68.78 \; \grayref{\pm1.0}$& N/A\\
\rowcolor{shadecolor}MT \cite{MT} & \grayref{[NeurIPS 2017]}& $79.23 \; \grayref{\pm0.6}$& $78.95 \; \grayref{\pm1.6}$& $77.25 \; \grayref{\pm0.7}$& $74.96 \; \grayref{\pm0.4}$& 3.73\\
CPS \cite{CPS} & \grayref{[CVPR 2021]}& $80.10 \; \grayref{\pm3.1}$& $77.43 \; \grayref{\pm2.9}$& $75.27 \; \grayref{\pm3.9}$& $73.67 \; \grayref{\pm4.6}$& 2.72\\
\rowcolor{shadecolor}RL \cite{RL} & \grayref{[MICCAI 2021]}& $80.21 \; \grayref{\pm1.0}$& $79.98 \; \grayref{\pm1.0}$& $79.05 \; \grayref{\pm1.4}$& $78.73 \; \grayref{\pm1.7}$& 6.71\\
STPP \cite{STPP} & \grayref{[CVPR 2022]}& $80.21 \; \grayref{\pm1.1}$& $79.63 \; \grayref{\pm2.4}$& $79.24 \; \grayref{\pm1.2}$& $74.40 \; \grayref{\pm2.3}$& 4.49\\
\rowcolor{shadecolor}PS-MT \cite{PSMT} & \grayref{[CVPR 2022]}& $78.86 \; \grayref{\pm1.0}$& $78.16 \; \grayref{\pm0.7}$& $77.34 \; \grayref{\pm0.8}$& $75.73 \; \grayref{\pm1.2}$& 3.87\\
MCF \cite{MCF} & \grayref{[CVPR 2023]}& $81.35 \; \grayref{\pm4.7}$& $77.53 \; \grayref{\pm3.9}$& $74.80 \; \grayref{\pm3.8}$& $71.19 \; \grayref{\pm1.1}$& 2.66\\
\rowcolor{shadecolor}BCP \cite{BCP} & \grayref{[CVPR 2023]}& $81.55 \; \grayref{\pm4.9}$& $79.89 \; \grayref{\pm6.0}$& $75.05 \; \grayref{\pm8.6}$& $74.73 \; \grayref{\pm6.4}$& 4.00\\
GPS \cite{GPS} & \grayref{[WACV 2024]}& $81.62 \; \grayref{\pm2.2}$& $80.60 \; \grayref{\pm3.5}$& $79.18 \; \grayref{\pm1.6}$& $73.04 \; \grayref{\pm2.5}$& 4.60\\
\specialrule{.02em}{.05em}{.05em}
\rowcolor{shadecolor}UniMatch \cite{WS} & \grayref{[CVPR 2023]}& $79.84 \; \grayref{\pm1.2}$& $79.36 \; \grayref{\pm0.1}$& $78.42 \; \grayref{\pm0.8}$& $77.34 \; \grayref{\pm1.0}$& 5.54\\
UniMatch + ENCORE && $81.47 \; \grayref{\pm1.3}$& $81.77 \; \grayref{\pm1.0}$& $80.93 \; \grayref{\pm1.6}$& $79.54 \; \grayref{\pm2.1}$& 7.62\\
\specialrule{.02em}{.05em}{.05em}
\rowcolor{shadecolor}Switch \cite{DualTeacher} & \grayref{[NeurIPS 2024]}& $78.80 \; \grayref{\pm1.2}$& $78.12 \; \grayref{\pm0.7}$& $77.35 \; \grayref{\pm0.6}$& $74.74 \; \grayref{\pm2.0}$& 3.76\\
Switch + ENCORE && $79.81 \; \grayref{\pm1.0}$& $80.04 \; \grayref{\pm0.6}$& $78.80 \; \grayref{\pm0.8}$& $76.59 \; \grayref{\pm1.8}$& 4.75\\
\specialrule{.02em}{.05em}{.05em}
\rowcolor{shadecolor}AD-MT \cite{AD-MT} & \grayref{[ECCV 2024]}& $80.63 \; \grayref{\pm1.2}$& $80.30 \; \grayref{\pm1.3}$& $79.63 \; \grayref{\pm0.9}$& $78.14 \; \grayref{\pm1.6}$& 6.52\\
AD-MT + ENCORE && $80.27 \; \grayref{\pm0.6}$& $80.97 \; \grayref{\pm1.0}$& $79.69 \; \grayref{\pm1.0}$& $78.14 \; \grayref{\pm1.1}$& 6.53\\
\specialrule{.12em}{.05em}{.05em} 
\end{tabular}
}
\end{table}

\begin{table}[th!]
\centering
\caption{Comparisons with state-of-the-art methods on \textbf{ACDC} test set (Network: UNet) based on dice coefficient (\%).}
\label{tab:ACDC}
\renewcommand{\arraystretch}{1.1}
\resizebox{1\columnwidth}{!}{%
\begin{tabular}{r*{1}@{\hspace{1.8pt}}l*{1}{>{\centering\arraybackslash}m{1.2cm}}*{5}{>{\centering\arraybackslash}m{1.4cm}}*{1}{>
{\centering\arraybackslash}m{2cm}}}
\specialrule{.12em}{.05em}{.05em}
Framework && 1/7 (20) & 1/14 (10)& 1/70 (2) & 1/140 (1)& Rel. Avg. \\
\specialrule{.12em}{.05em}{.05em} 
Supervised && 86.38& 84.24& 34.23&28.38& N/A \\
\specialrule{.02em}{.05em}{.05em}
\rowcolor{shadecolor}UniMatch \cite{WS}& \grayref{[CVPR 2023]}& 90.07& 88.54 & 80.98 & 78.41 & 26.19 \\
UniMatch + ENCORE && 90.07 & 88.68 & 85.81 & 84.88 & 29.05\\\specialrule{.02em}{.05em}{.05em}
\rowcolor{shadecolor}AD-MT \cite{AD-MT} & \grayref{[ECCV 2024]}&  89.71 & 88.61 & 80.65  & 75.06  & 25.20 \\
AD-MT + ENCORE && 89.54  & 88.21  & 85.61  & 80.48  & 27.65 \\
\specialrule{.12em}{.05em}{.05em} 
\end{tabular}%
}
\end{table}

\begin{table}[tb!]
\centering
\caption{Comparisons with state-of-the-art methods on \textbf{LA} test set (Network: VNet) based on dice coefficient (\%).}
\label{tab:LA}
\renewcommand{\arraystretch}{1.1}
\resizebox{1\columnwidth}{!}{
\begin{tabular}{r@{\hspace{1.8pt}}l >{\centering\arraybackslash}m{1.2cm} *{4}{>{\centering\arraybackslash}m{1.4cm}} >{\centering\arraybackslash}m{1cm}}
\specialrule{.12em}{.05em}{.05em}
Framework && 1/10 (8)& 1/20 (4) &  1/26 (3) & 1/40 (2) & Rel. Avg. \\
\specialrule{.12em}{.05em}{.05em}
supervised&& 84.04& 60.86& 65.15& 33.79& N/A\\
\specialrule{.02em}{.05em}{.05em}
\rowcolor{shadecolor}UniMatch \cite{WS}& \grayref{[CVPR 2023]}& 89.52 & 87.53 & 85.67 & 84.24 & 25.78 \\
UniMatch + ENCORE2 &&89.76 & 86.94 & 87.75 & 86.18 & 26.70 \\

\specialrule{.02em}{.05em}{.05em}
    
\rowcolor{shadecolor}Switch\cite{DualTeacher}&\grayref{[NeurIPS 2024]}& 89.70& 87.85& 88.84& 86.60& 27.29 \\

Switch + ENCORE2 &&89.62&88.66&88.81&86.73& 27.50\\

\specialrule{.02em}{.05em}{.05em}
\rowcolor{shadecolor} AD-MT \cite{AD-MT}& \grayref{[ECCV 2024]} &90.01  & 89.76 & 88.89& 83.21 & 27.01\\

AD-MT + ENCORE2 &&89.46 & 89.42 & 89.25 & 89.13 & 8.36 \\

\specialrule{.12em}{.05em}{.05em}
\end{tabular}
}
\end{table}

\begin{table}[th!]
\centering
\caption{Ablation Study on one fold of \textbf{Cataract-1K} dataset with Switch framework based on Dice coefficient (\%). The first and the second lines for each data split correspond to Supervised training and Switch, respectively.}
\label{tab:ablation-Cat1K}
\renewcommand{\arraystretch}{1.1}
\resizebox{1\columnwidth}{!}{%
\begin{tabular}{>{\centering\arraybackslash}m{1.5cm}*{16}{>{\centering\arraybackslash}m{1.3cm}}}
\specialrule{.12em}{.05em}{.05em}
&&&&\multicolumn{5}{c}{Dice (\%)}\\
\cmidrule(lr){5-9}
 & Semi & CAC & ACT & Pupil & Iris & Lens & Instrument & Avg  \\
\specialrule{.12em}{.05em}{.05em}
\multirow{4}{*}{\centering \textbf{1/8 (242)}}&\XSolidBrush & \XSolidBrush & \XSolidBrush & 83.67 & 67.86 & 80.44 & 72.76 & 76.18 \\
&\CheckmarkBold & \XSolidBrush & \XSolidBrush& 83.75 & 70.02 & 76.79 & 76.67 & 76.81 \\
&\CheckmarkBold & \XSolidBrush & \CheckmarkBold& 90.93 & 83.48 & 87.42 & 82.58 & 86.11 \\
&\CheckmarkBold & \CheckmarkBold & \XSolidBrush & 90.53 & 82.83 & 87.13 & 82.57 & 85.77\\
&\CheckmarkBold & \CheckmarkBold & \CheckmarkBold & \textbf{93.61} & \textbf{85.59} & \textbf{91.76} & \textbf{87.51} & \textbf{89.62} \\

\specialrule{.12em}{.05em}{.05em}
\multirow{4}{*}{\centering \textbf{1/26 (61)}}&\XSolidBrush & \XSolidBrush & \XSolidBrush & 76.65 & 62.48 & 70.08 & 46.91 & 64.03\\
&\CheckmarkBold & \XSolidBrush & \XSolidBrush& 84.28 & 61.83 & 64.92 & 52.94 & 66.01 \\
&\CheckmarkBold & \XSolidBrush & \CheckmarkBold& 87.31 & 73.72 & 80.27 & 68.51 & 77.45 \\
&\CheckmarkBold & \CheckmarkBold & \XSolidBrush & 87.22 & 72.76 & 78.67 & 65.57 & 76.05 \\
&\CheckmarkBold & \CheckmarkBold & \CheckmarkBold & \textbf{92.61} & \textbf{86.78} & \textbf{83.74} & \textbf{84.24} & \textbf{86.84} \\
\specialrule{.12em}{.05em}{.05em}
\end{tabular}
}
\end{table}

\begin{table}[th!]
\centering
\caption{Ablation Study on \textbf{ACDC} and \textbf{LA} datasets with AD-MT framework based on different metrics (\%). The first and second lines for each dataset correspond to Supervised training and AD-MT, respectively.}
\label{tab:ablation}
\renewcommand{\arraystretch}{1.1}
\resizebox{0.44\textwidth}{!}{%
\begin{tabular}{>{\centering\arraybackslash}m{0.8cm} >{\centering\arraybackslash}m{0.8cm} *{6}{@{} >{\centering\arraybackslash}m{1.5cm} >{\centering\arraybackslash}m{1.5cm} @{}}}
\specialrule{.12em}{.05em}{.05em}
& & \multicolumn{9}{c}{ACDC} \\
\cmidrule(lr){4-10}
& & \multicolumn{4}{c}{2} & \multicolumn{3}{c}{1} \\
\cmidrule(lr){4-6} \cmidrule(lr){7-9} 
Semi & CAC & ACT & Dice $\uparrow$ & 95HD $\downarrow$ & ASD $\downarrow$ & Dice $\uparrow$ & 95HD $\downarrow$ & ASD $\downarrow$ \\
\specialrule{.12em}{.05em}{.05em}
\XSolidBrush & \XSolidBrush & \XSolidBrush & 34.23 & 52.18 & 23.91 & 28.38 & 65.41 & 31.86\\
\CheckmarkBold & \XSolidBrush & \XSolidBrush & 80.65 & 6.22 & 1.67 & 75.06 & 20.36 & 7.87 \\
\CheckmarkBold & \XSolidBrush & \CheckmarkBold & 84.66 & 5.72 & \textbf{1.40} & \textbf{80.60} & \textbf{3.19} & \textbf{0.99} \\

\CheckmarkBold & \CheckmarkBold & \CheckmarkBold & \textbf{85.61} & \textbf{5.45} & 1.47 & 80.48 & 4.76 & 1.43 \\
\end{tabular}
}
\resizebox{0.44\textwidth}{!}{%
\begin{tabular}{>{\centering\arraybackslash}m{0.8cm} >{\centering\arraybackslash}m{0.8cm} *{6}{@{} >{\centering\arraybackslash}m{1.5cm} >{\centering\arraybackslash}m{1.5cm} @{}}}
\specialrule{.12em}{.05em}{.05em}
&&\multicolumn{9}{c}{LA}\\
\cmidrule(lr){4-10}
& & \multicolumn{4}{c}{3} & \multicolumn{3}{c}{2} \\
\cmidrule(lr){4-6} \cmidrule(lr){7-9} 
Semi & CAC & ACT & Dice $\uparrow$ & 95HD $\downarrow$ & ASD $\downarrow$ & Dice $\uparrow$ & 95HD $\downarrow$ & ASD $\downarrow$ \\
\specialrule{.12em}{.05em}{.05em}
\XSolidBrush & \XSolidBrush & \XSolidBrush & 65.15& 32.13 & 3.59 & 33.79& 59.82 & 11.45\\
\CheckmarkBold & \XSolidBrush & \XSolidBrush & 88.89& 7.22 & 2.00 & 83.21 & 11.37 & 2.79\\
\CheckmarkBold & \XSolidBrush & \CheckmarkBold & 88.58 & 7.87 & 1.86 & 87.64 & 8.23 & 2.24 \\
\CheckmarkBold & \CheckmarkBold & \CheckmarkBold & \textbf{89.25} &\textbf{7.03} & \textbf{1.77} & \textbf{89.13} & \textbf{7.11} & \textbf{1.96}\\
\specialrule{.12em}{.05em}{.05em}
\end{tabular}}
\end{table}

Tables~\ref{tab:Cat1k}, \ref{tab:MRI}, \ref{tab:EndoVis}, \ref{tab:ACDC}, and \ref{tab:LA} compare the performance of our proposed feedback-driven teacher (ENCORE) against supervised learning and multiple semi-supervised baselines. The fractions indicate the percentage of labeled data used for training. The relative average Dice score is computed as the average relative improvement over the supervised baseline across all evaluated splits. For the Prostate MRI and EndoVis datasets, this relative average also corresponds to the extended evaluations provided in the supplementary material.  

For multi-class segmentation in Cataract-1K (Table~\ref{tab:Cat1k}), incorporating our dynamic thresholding method (ENCORE) led to substantial improvements over state-of-the-art methods with static thresholds. Notably, applying ENCORE to UniMatch with just 1/26 labeled data (61 images) yielded higher performance than UniMatch trained on 1/4 labeled data (433 images) alone. Additionally, integrating ENCORE with the three baselines significantly reduced the standard deviation of Dice scores across four folds in the Cataract-1K dataset. Furthermore, adding ENCORE to UniMatch, Switch, and AD-MT consistently positioned these methods as the top-performing approaches against all baselines.  

For the Prostate MRI dataset (Table~\ref{tab:MRI}), ENCORE demonstrated superior performance even in extreme low-data regimes. With only 1/14 labeled data (279 images), ENCORE outperformed semi-supervised training on 1/4 labeled data (872 images) with both UniMatch and Switch, and also surpassed supervised training with 1/2 labeled data (1673 images). The overall performance remained higher across all evaluated methods, primarily due to ENCORE’s effectiveness in low-data regimes (e.g., 78.79\% vs. 70.04\% for Switch). Similar trends were observed on the EndoVis dataset, where UniMatch + ENCORE achieved the highest relative average improvement compared to state-of-the-art methods.  

\begin{figure*}[t!]
  \centering
  % Define a fixed height for all subfigures
  \newcommand{\imgheight}{7.3cm} % Adjust this value as needed

  % First row (3 subfigures)
  \begin{subfigure}[t]{\textwidth}
    \centering
    \includegraphics[height=\imgheight,keepaspectratio]{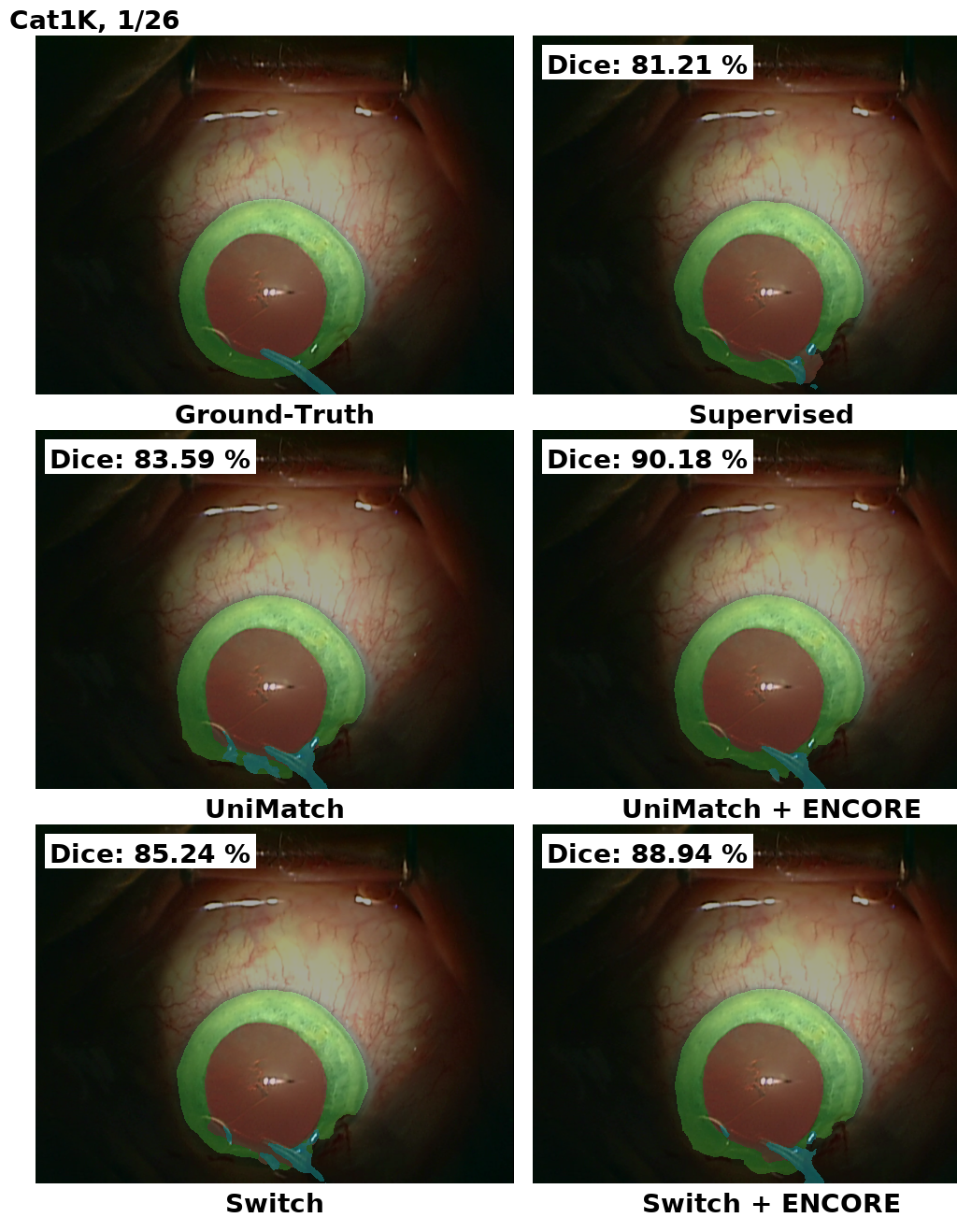}
    \includegraphics[height=\imgheight,keepaspectratio]{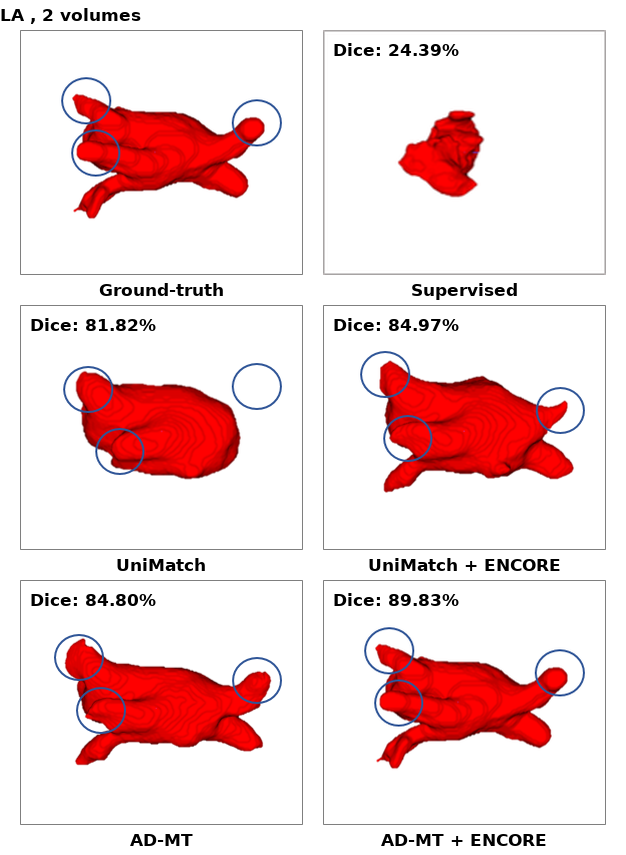}
    \includegraphics[height=\imgheight,keepaspectratio]{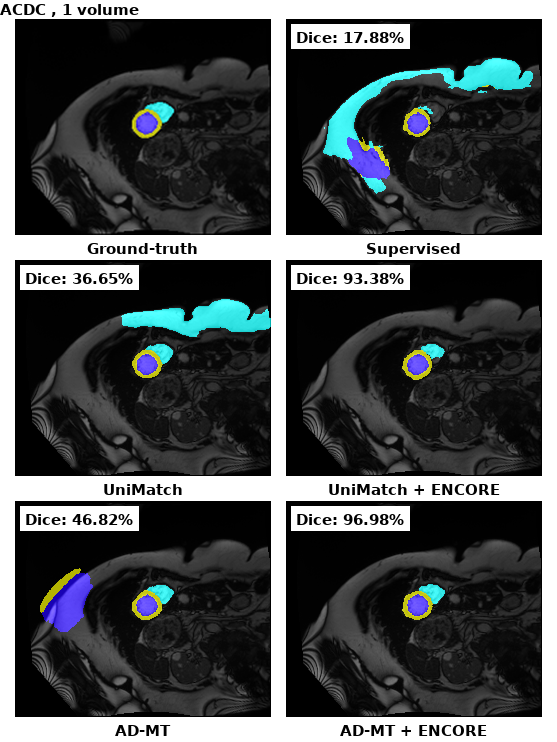}
  \end{subfigure}

  \caption{Qualitative comparisons of state-of-the-art methods with and without ENCORE on the Cataract-1K, LA, and ACDC datasets.}
  \label{fig:qualitative}
\end{figure*}

For multi-class segmentation in the ACDC dataset (Table~\ref{tab:ACDC}), ENCORE further demonstrated its effectiveness in low-data settings. UniMatch + ENCORE trained with only one labeled volume outperformed supervised training with 10 labeled volumes. As expected, performance gains with ENCORE were most pronounced in extreme low-data settings (one and two labeled volumes); however, improvements remained consistently high across different semi-supervised learning frameworks.  

Moreover, these gains were consistent across different network architectures: DeepLabV3+ for Tables~\ref{tab:Cat1k}, \ref{tab:MRI}, and \ref{tab:EndoVis}, UNet for Table~\ref{tab:ACDC}, and VNet for Table~\ref{tab:LA}.  
Overall, the combination of ENCORE with state-of-the-art methods consistently emerged as the top-performing approach across all reported datasets.  

\noindent\textbf{Ablation Study. }  
Since ENCORE consists of two key components—Class-Aware Confidence Calibration (CAC) and Adaptive Confidence Thresholding (ACT)—we conducted an ablation study to assess their individual contributions to segmentation performance.  
Table~\ref{tab:ablation-Cat1K} evaluates the impact of CAC and ACT on per-class segmentation performance in the Cataract-1K dataset using the Switch framework. While Switch alone improved the average Dice score over supervised learning, a per-class analysis revealed performance degradation for challenging classes, particularly Lens in 1/8 labeled data and Iris and Lens in 1/26 labeled data. In contrast, integrating either CAC or ACT independently improved segmentation performance across all classes. Moreover, combining both CAC and ACT yielded a significantly larger improvement compared to using each module in isolation.  
Table~\ref{tab:ablation} presents the results of the ablation study on the ACDC and LA datasets, further reinforcing the contribution of our proposed modules in improving segmentation performance when integrated into the AD-MT framework.

Figure~\ref{fig:qualitative} presents qualitative comparisons between different baselines with and without ENCORE. Notably, for the LA dataset, integrating ENCORE into both UniMatch and AD-MT enhances the retrieval of important structural features in volumetric segmentation.  

\begin{figure}[!t]
    \centering
    \includegraphics[width=0.5\textwidth]{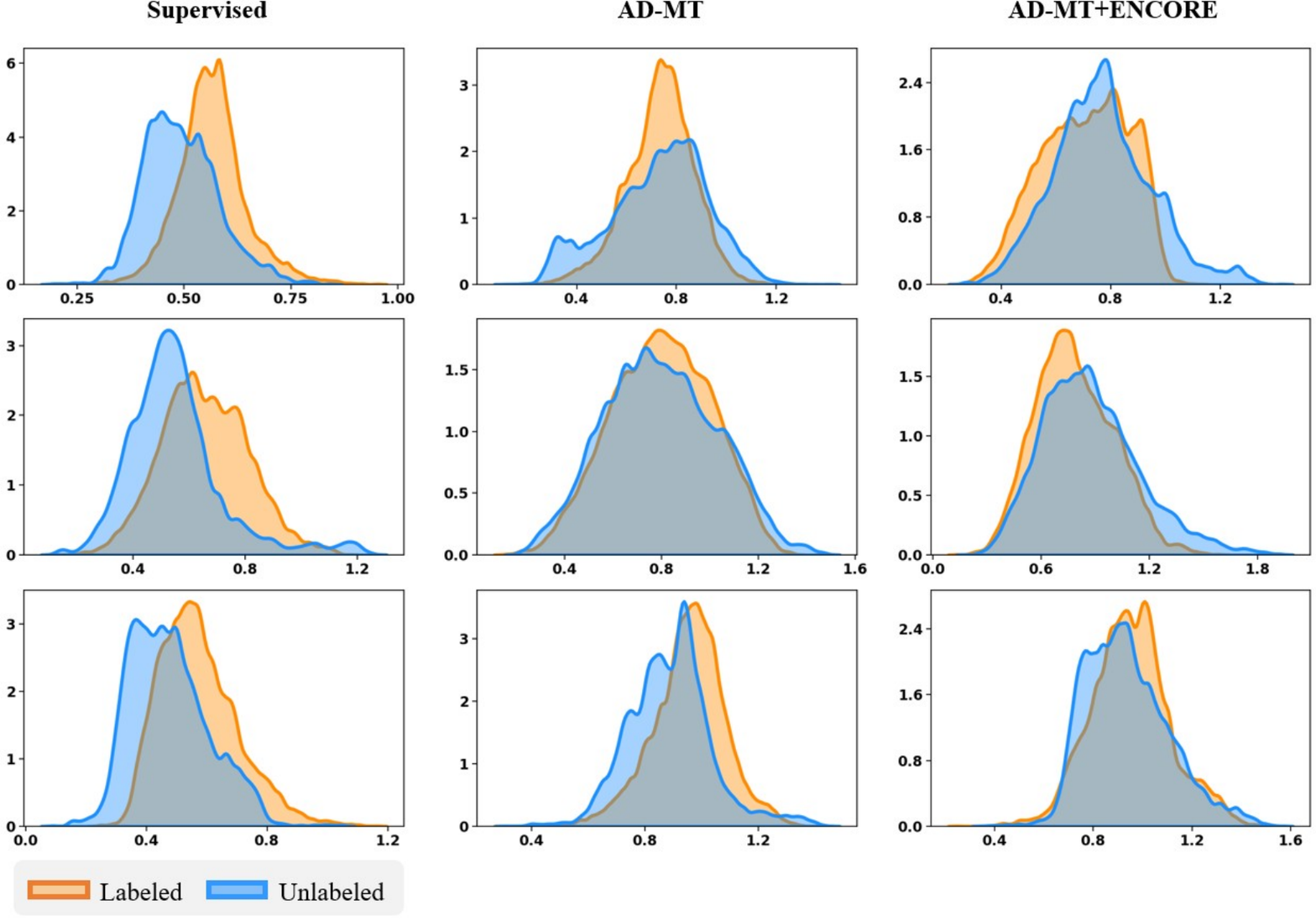}
    \caption{Kernel density estimates from various methods trained on 1 labeled ACDC dataset. From top to bottom, the plots display features corresponding to the right ventricle, myocardium, and left ventricle classes.}
    \label{fig:kde}
\end{figure}

Figure~\ref{fig:kde} compares the Kernel Density Estimation (KDE) of the AD-MT method with and without ENCORE across three classes in the ACDC dataset. The figure illustrates that while AD-MT alone better aligns the data distributions of labeled and unlabeled samples compared to supervised training, the addition of ENCORE further bridges the distribution gap, leading to improved feature consistency.  

For additional discussions, experiments, and visual comparisons, we refer reviewers to the supplementary material.

\section{Conclusion}
\label{sec:conclusion}

In this paper, we introduced \textit{Ensemble-of-Confidence Reinforcement (ENCORE)}, a dynamic feedback-driven pseudo-label thresholding strategy for semi-supervised semantic segmentation. Unlike conventional methods that rely on a fixed confidence threshold, ENCORE adaptively refines class-wise pseudo-label selection by leveraging a feedback mechanism. Our approach dynamically adjusts thresholds based on the student network’s performance, ensuring that informative pseudo-labels are retained while unreliable ones are filtered out. By continuously recalibrating thresholds throughout training, ENCORE mitigates overconfident noise and prevents the premature exclusion of meaningful pseudo-labels, thereby improving segmentation performance across different structures.
Extensive experiments across five datasets, three network architectures, and multiple labeled data fractions, benchmarked against state-of-the-art baselines, validate the effectiveness of our method. The proposed feedback-driven pseudo-label reliability assessment method can be seamlessly integrated with pseudo-supervision frameworks, further enhancing semantic segmentation performance. 
In future work, we aim to extend ENCORE to broader tasks such as classification and action recognition, exploring its applicability beyond semantic segmentation.

{
    \small
    \balance
    \bibliographystyle{ieeenat_fullname}
    \bibliography{main}
}

\end{document}

\maketitle

\section{Discussion}
\label{sec: Discussion}

In this work, we shift our focus from conventional pseudo-label refinement to the critical issue of \textit{confidence disparities} across different classes. Pseudo-labeling methods typically rely on a fixed confidence threshold to filter unreliable predictions, yet confidence scores vary significantly across different structures. In medical image segmentation, for instance, large anatomical regions are often predicted with high confidence, whereas smaller or more complex structures exhibit lower confidence, leading to systematic errors in pseudo-label selection. This issue is well-documented in semi-supervised learning, where highly confident pseudo-labels tend to correspond to more frequently occurring patterns~\cite{liu2022acpl}. As a result, using a global confidence threshold disproportionately favors high-confidence predictions while filtering out numerous correct but lower-confidence pseudo-labels. This can lead to suboptimal learning, as valuable training signals from certain structures are excluded, reinforcing model biases toward overrepresented patterns.

Furthermore, confidence levels are not only structure-dependent but also vary across different classes with differing complexities and appearances. Some classes naturally achieve higher confidence early in training, while others remain uncertain throughout multiple training stages~\cite{chen2023class}. This per-class confidence heterogeneity renders a static threshold ineffective, as a single threshold may be too strict for challenging classes, discarding valuable pseudo-labels, or too lenient for more easily learned classes, introducing excessive noise. Determining class-specific thresholds a priori is infeasible, particularly when the true confidence distribution varies throughout training. Consequently, fixed-threshold methods often discard informative samples from harder-to-learn structures, limiting pseudo-label diversity and overall model generalization. A more adaptive approach is needed—one that adjusts thresholding dynamically rather than enforcing a uniform criterion.

To address these challenges, we propose a \textit{feedback-driven dynamic thresholding} approach for pseudo-label selection. Instead of a fixed confidence cutoff, our method continuously refines class-wise thresholds based on model feedback from both labeled and unlabeled data. The model adaptively adjusts the threshold per class by monitoring prediction reliability, ensuring that informative pseudo-labels are retained while noisy predictions are filtered appropriately. This eliminates the need for manual threshold tuning while maintaining a balance between inclusiveness and precision in pseudo-label selection.

Unlike traditional fixed-threshold methods that risk excluding valuable training samples or incorporating unreliable pseudo-labels, our feedback-driven strategy dynamically recalibrates confidence thresholds throughout training. By preventing premature exclusion of meaningful pseudo-labels while controlling overconfident noise, our method ensures that each class contributes optimally to the learning process. As a result, the proposed approach leads to a more diverse, high-quality set of pseudo-labels, improving segmentation performance across different structures.
\begin{figure}[!t]
    \centering
    \includegraphics[width=0.45\textwidth]{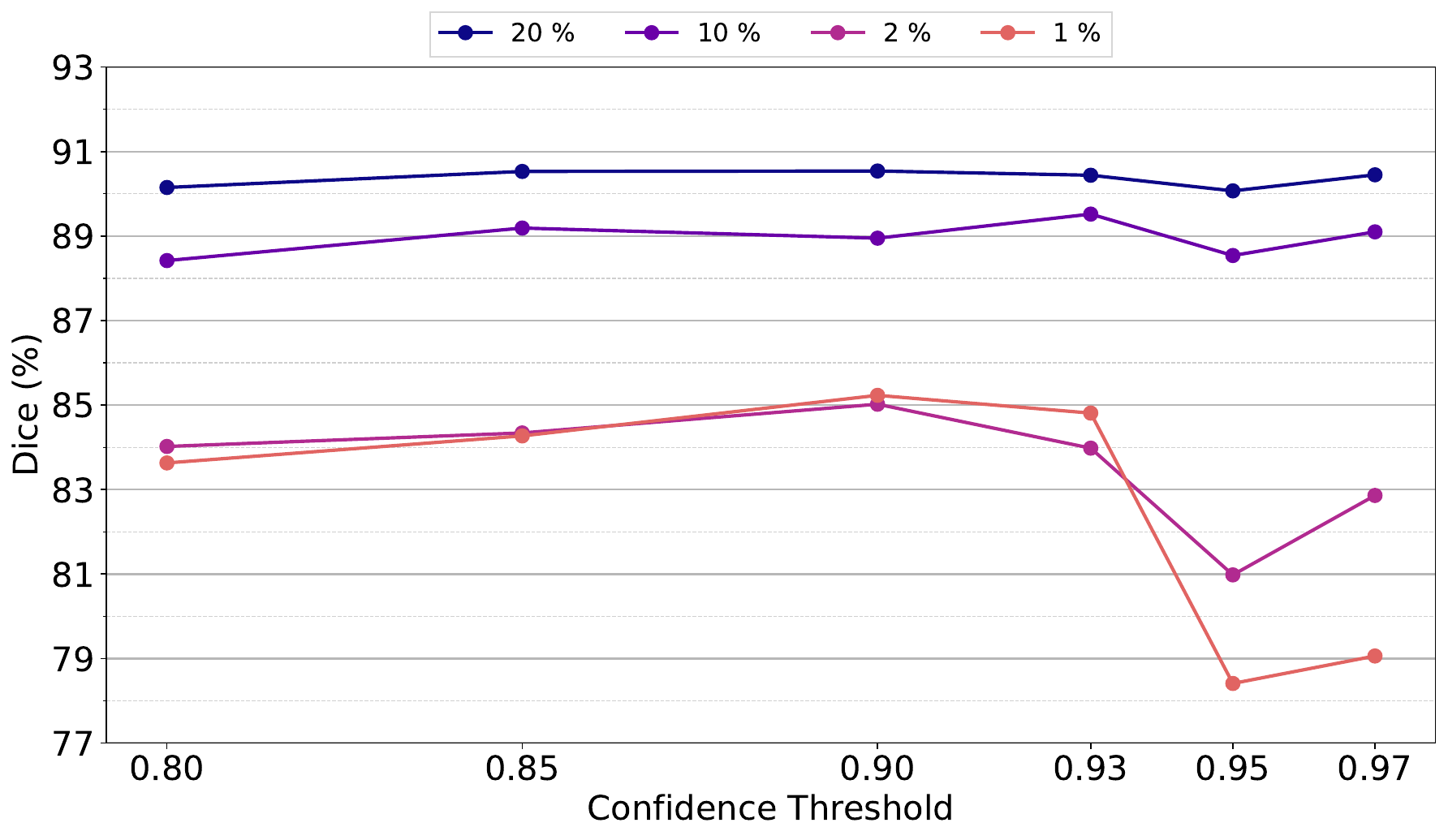}
    \caption{Comparison of dice scores across thresholds for various labeled data percentages on ACDC dataset with UniMatch \cite{WS}.}
    \label{fig:thresholdvariation_weakstrong}
\end{figure}

\section{Experimental Results}
\label{sec:experimental results}
This section provides all supplementary results and visualizations for the main paper.

\begin{table*}[th!]
\centering
\caption{Comparisons with state-of-the-art methods on \textbf{Prostate MRI} test set based on dice coefficient ($\%$).}
\label{tab:MRI_sup}
\renewcommand{\arraystretch}{1.1}
\resizebox{0.9\textwidth}{!}{%
\begin{tabular}{r*{1}@{\hspace{2pt}}l*{1}{>{\centering\arraybackslash}m{2cm}}*{6}{>{\centering\arraybackslash}m{2cm}}*{1}{>{\centering\arraybackslash}m{1cm}}}
\specialrule{.12em}{.05em}{.05em}
Framework &&1/2 (1673)&1/4 (872)&1/8 (444)&1/14 (279)&1/25 (156)&1/50 (64)& Rel. Avg. \\
\specialrule{.12em}{.05em}{.05em} 
Supervised && $77.81 \; \grayref{\pm3.1}$& $74.46 \; \grayref{\pm3.1}$& $72.37 \; \grayref{\pm1.4}$& $65.05 \; \grayref{\pm4.9}$& $58.99 \; \grayref{\pm5.7}$& $53.51 \; \grayref{\pm1.7}$& N/A\\
MT \cite{MT} & \grayref{[NeurIPS 2017]}& \rowcolor{shadecolor}$82.47 \; \grayref{\pm1.3}$& $81.73 \; \grayref{\pm2.0}$& $79.71 \; \grayref{\pm1.7}$& $75.79 \; \grayref{\pm3.1}$& $69.02 \; \grayref{\pm5.1}$& $64.01 \; \grayref{\pm4.0}$& 8.42\\
CPS \cite{CPS} & \grayref{[CVPR 2021]}& $78.48 \; \grayref{\pm2.2}$& $76.94 \; \grayref{\pm4.4}$& $72.72 \; \grayref{\pm2.7}$& $64.78 \; \grayref{\pm4.5}$& $62.80 \; \grayref{\pm4.5}$& $57.50 \; \grayref{\pm1.3}$& 1.84\\
RL \cite{RL} & \grayref{[MICCAI 2021]}& \rowcolor{shadecolor}$79.87 \; \grayref{\pm1.7}$& $78.48 \; \grayref{\pm2.8}$& $77.97 \; \grayref{\pm2.7}$& $72.28 \; \grayref{\pm3.6}$& $69.92 \; \grayref{\pm2.4}$& $66.42 \; \grayref{\pm4.5}$& 7.12\\
ST++ \cite{STPP} & \grayref{[CVPR 2022]}& $79.44 \; \grayref{\pm1.3}$& $76.47 \; \grayref{\pm2.7}$& $75.36 \; \grayref{\pm1.7}$& $69.68 \; \grayref{\pm7.5}$& $60.77 \; \grayref{\pm8.4}$& $55.72 \; \grayref{\pm4.1}$& 2.54\\
PS-MT \cite{PSMT} & \grayref{[CVPR 2022]}& \rowcolor{shadecolor}$79.83 \; \grayref{\pm1.0}$& $77.41 \; \grayref{\pm3.8}$& $74.89 \; \grayref{\pm2.2}$& $69.42 \; \grayref{\pm4.6}$& $67.24 \; \grayref{\pm3.0}$& $63.20 \; \grayref{\pm2.6}$& 4.97\\
MCF \cite{MCF} & \grayref{[CVPR 2023]}& $76.96 \; \grayref{\pm1.6}$& $73.57 \; \grayref{\pm3.9}$& $73.51 \; \grayref{\pm1.8}$& $71.94 \; \grayref{\pm4.7}$& $67.54 \; \grayref{\pm1.9}$& $63.09 \; \grayref{\pm3.7}$& 4.07\\
BCP \cite{BCP} & \grayref{[CVPR 2023]}& \rowcolor{shadecolor}$80.16 \; \grayref{\pm1.4}$& $78.95 \; \grayref{\pm1.1}$& $74.65 \; \grayref{\pm2.4}$& $69.74 \; \grayref{\pm0.8}$& $65.49 \; \grayref{\pm1.9}$& $56.40 \; \grayref{\pm3.6}$& 3.87\\
GPS \cite{GPS} & \grayref{[WACV 2024]}& $83.94 \; \grayref{\pm0.9}$& $83.49  \; \grayref{\pm1.9}$& $81.17  \; \grayref{\pm1.9}$& $79.02  \; \grayref{\pm2.7}$& $72.93  \; \grayref{\pm7.4}$& $65.99 \; \grayref{\pm4.8}$& 10.73\\
\specialrule{.12em}{.05em}{.05em} 
UniMatch \cite{WS} & \grayref{[CVPR 2023]}& $79.96 \; \grayref{\pm1.0}$& $77.94 \; \grayref{\pm2.1}$& $74.42 \; \grayref{\pm1.9}$& $73.86 \; \grayref{\pm2.3}$& $66.64 \; \grayref{\pm5.3}$& $58.31 \; \grayref{\pm5.0}$& 4.82\\
UniMatch + ENCORE && \rowcolor{shadecolor} $\textbf{82.45} \; \grayref{\pm0.8}$& $\textbf{81.59} \; \grayref{\pm1.8}$& $\textbf{78.85} \; \grayref{\pm4.5}$& $\textbf{78.65} \; \grayref{\pm2.2}$& $\textbf{73.30} \; \grayref{\pm3.2}$& $\textbf{66.82} \; \grayref{\pm3.2}$& \textbf{9.91}\\
\specialrule{.12em}{.05em}{.05em} 
Switch \cite{DualTeacher} & \grayref{[NeurIPS 2024]}& $79.45 \; \grayref{\pm0.4}$& $77.66 \; \grayref{\pm3.6}$& $75.72 \; \grayref{\pm2.8}$& $70.04 \; \grayref{\pm4.1}$& $66.11 \; \grayref{\pm4.3}$& $61.94 \; \grayref{\pm2.5}$& 4.79\\
Switch + ENCORE && $\textbf{83.98} \; \grayref{\pm0.8}$& $\textbf{84.24} \; \grayref{\pm1.6}$& $\textbf{78.96} \; \grayref{\pm5.1}$& $\textbf{77.89} \; \grayref{\pm4.7}$& $\textbf{70.68} \; \grayref{\pm7.4}$& $\textbf{65.29} \; \grayref{\pm4.5}$& \textbf{9.81}\\
\specialrule{.12em}{.05em}{.05em} 
AD-MT \cite{AD-MT} & \grayref{[ECCV 2024]}& $82.21 \; \grayref{\pm1.3}$& $80.03 \; \grayref{\pm2.3}$& $78.26 \; \grayref{\pm2.0}$& $75.00 \; \grayref{\pm3.5}$& $\textbf{68.43} \; \grayref{\pm5.0}$& $64.29 \; \grayref{\pm3.6}$& 7.67\\
AD-MT + ENCORE && \rowcolor{shadecolor} $\textbf{83.08} \; \grayref{\pm0.9}$& $\textbf{80.49} \; \grayref{\pm2.1}$& $\textbf{80.21} \; \grayref{\pm1.0}$& $\textbf{76.08} \; \grayref{\pm2.4}$& $67.99 \; \grayref{\pm5.7}$& $\textbf{64.86} \; \grayref{\pm4.5}$& \textbf{8.42}\\
\specialrule{.12em}{.05em}{.05em}

\end{tabular}
}
\end{table*}
\begin{table*}[th!]
\centering
\caption{Comparisons with state-of-the-art methods on \textbf{EndoVis} test set based on dice coefficient ($\%$).}
\label{tab:EndoVis_sup}
\renewcommand{\arraystretch}{1.1}
\resizebox{0.85\textwidth}{!}{%
\begin{tabular}{r*{1}@{\hspace{2pt}}l*{1}{>{\centering\arraybackslash}m{2cm}}*{5}{>{\centering\arraybackslash}m{2cm}}*{1}{>{\centering\arraybackslash}m{1cm}}}
\specialrule{.12em}{.05em}{.05em}
Framework &&1/2 (1764)&1/4 (882)&1/8 (441)&1/16 (221)&1/32 (111)& Rel. Avg. \\
\specialrule{.12em}{.05em}{.05em} 
Supervised && $78.46 \; \grayref{\pm1.1}$& $75.40 \; \grayref{\pm1.2}$& $72.28 \; \grayref{\pm0.8}$& $68.78 \; \grayref{\pm1.0}$& $67.70 \; \grayref{\pm1.0}$& N/A\\
MT \cite{MT} & \grayref{[NeurIPS 2017]}& \rowcolor{shadecolor}$79.23 \; \grayref{\pm0.6}$& $78.95 \; \grayref{\pm1.6}$& $77.25 \; \grayref{\pm0.7}$& $74.96 \; \grayref{\pm0.4}$& $70.88 \; \grayref{\pm1.8}$& 3.73\\
CPS \cite{CPS} & \grayref{[CVPR 2021]}& $80.10 \; \grayref{\pm3.1}$& $77.43 \; \grayref{\pm2.9}$& $75.27 \; \grayref{\pm3.9}$& $73.67 \; \grayref{\pm4.6}$& $69.75 \; \grayref{\pm4.4}$& 2.72\\
RL \cite{RL} & \grayref{[MICCAI 2021]}& \rowcolor{shadecolor}$80.21 \; \grayref{\pm1.0}$& $79.98 \; \grayref{\pm1.0}$& $79.05 \; \grayref{\pm1.4}$& $78.73 \; \grayref{\pm1.7}$& $78.19 \; \grayref{\pm1.1}$& 6.71\\
STPP \cite{STPP} & \grayref{[CVPR 2022]}& $80.21 \; \grayref{\pm1.1}$& $79.63 \; \grayref{\pm2.4}$& $79.24 \; \grayref{\pm1.2}$& $74.40 \; \grayref{\pm2.3}$& $71.59 \; \grayref{\pm0.6}$& 4.49\\
PS-MT \cite{PSMT} & \grayref{[CVPR 2022]}& \rowcolor{shadecolor}$78.86 \; \grayref{\pm1.0}$& $78.16 \; \grayref{\pm0.7}$& $77.34 \; \grayref{\pm0.8}$& $75.73 \; \grayref{\pm1.2}$& $71.88 \; \grayref{\pm1.4}$& 3.87\\
MCF \cite{MCF} & \grayref{[CVPR 2023]}& $81.35 \; \grayref{\pm4.7}$& $77.53 \; \grayref{\pm3.9}$& $74.80 \; \grayref{\pm3.8}$& $71.19 \; \grayref{\pm1.1}$& $71.07 \; \grayref{\pm1.9}$& 2.66\\
BCP \cite{BCP} & \grayref{[CVPR 2023]}& \rowcolor{shadecolor}$81.55 \; \grayref{\pm4.9}$& $79.89 \; \grayref{\pm6.0}$& $75.05 \; \grayref{\pm8.6}$& $74.73 \; \grayref{\pm6.4}$& $71.40 \; \grayref{\pm3.0}$& 4.00\\
GPS \cite{GPS} & \grayref{[WACV 2024]}& $81.62 \; \grayref{\pm2.2}$& $80.60 \; \grayref{\pm3.5}$& $79.18 \; \grayref{\pm1.6}$& $73.04 \; \grayref{\pm2.5}$& $71.16 \; \grayref{\pm3.6}$& 4.60\\
\specialrule{.12em}{.05em}{.05em} 
UniMatch \cite{WS} & \grayref{[CVPR 2023]}& $79.84 \; \grayref{\pm1.2}$& $79.36 \; \grayref{\pm0.1}$& $78.42 \; \grayref{\pm0.8}$& $77.34 \; \grayref{\pm1.0}$& $75.34 \; \grayref{\pm2.1}$& 5.54\\
UniMatch + ENCORE && \rowcolor{shadecolor}$\textbf{81.47} \; \grayref{\pm1.3}$& $\textbf{81.77} \; \grayref{\pm1.0}$& $\textbf{80.93} \; \grayref{\pm1.6}$& $\textbf{79.54} \; \grayref{\pm2.1}$& $\textbf{77.03} \; \grayref{\pm1.1}$& \textbf{7.62}\\
\specialrule{.12em}{.05em}{.05em} 
Switch \cite{DualTeacher} & \grayref{[NeurIPS 2024]}& \rowcolor{shadecolor}$78.80 \; \grayref{\pm1.2}$& $78.12 \; \grayref{\pm0.7}$& $77.35 \; \grayref{\pm0.6}$& $74.74 \; \grayref{\pm2.0}$& $\textbf{72.41} \; \grayref{\pm1.4}$& 3.76\\
Switch + ENCORE && $\textbf{79.81} \; \grayref{\pm1.0}$& $\textbf{80.04} \; \grayref{\pm0.6}$& $\textbf{78.80} \; \grayref{\pm0.8}$& $\textbf{76.59} \; \grayref{\pm1.8}$& $71.13 \; \grayref{\pm0.9}$& \textbf{4.75}\\
\specialrule{.12em}{.05em}{.05em} 
AD-MT \cite{AD-MT} & \grayref{[ECCV 2024]}& \rowcolor{shadecolor}$\textbf{80.63} \; \grayref{\pm1.2}$& $80.30 \; \grayref{\pm1.3}$& $79.63 \; \grayref{\pm0.9}$& $78.14 \; \grayref{\pm1.6}$& $\textbf{76.50} \; \grayref{\pm0.6}$& 6.52\\
AD-MT + ENCORE && \rowcolor{shadecolor}$80.27 \; \grayref{\pm0.6}$& $\textbf{80.97} \; \grayref{\pm1.0}$& $\textbf{79.69} \; \grayref{\pm1.0}$& $78.14 \; \grayref{\pm1.1}$& $76.21 \; \grayref{\pm0.8}$& \textbf{6.53}\\
\specialrule{.12em}{.05em}{.05em} 
\end{tabular}
}
\end{table*}
% This is for Supplementary
\begin{table}[th!]
\centering
\caption{Comparisons with state-of-the-art methods on \textbf{ACDC} test set based on dice, 95HD, and ASD coefficients (\%).}
\label{tab:ACDC_sup}
\renewcommand{\arraystretch}{1.1}
\resizebox{1\columnwidth}{!}{%
\begin{tabular}{r*{1}@{\hspace{1.8pt}}l*{1}{>{\centering\arraybackslash}m{1.5cm}}*{5}{>{\centering\arraybackslash}m{1.5cm}}*{1}{>
{\centering\arraybackslash}m{0.5cm}}}
\specialrule{.12em}{.05em}{.05em}
Framework && Metric & 1/7 (20)& 1/14 (10)& 1/70 (2) & 1/140 (1)& Rel. Avg. \\
\specialrule{.12em}{.05em}{.05em} 
 && Dice &86.38& 84.24& 34.23&28.38& N/A \\
Supervised && 95HD& 6.47& 10.34& 52.18&65.41 & \\
 && ASD & 2.15& 2.80& 34.23&31.86 & \\
\specialrule{0.05em}{.05em}{.05em} 

&&Dice&\rowcolor{shadecolor} \textbf{90.07}& 88.54 & 80.98 & 78.41 & 26.19\\
UniMatch \cite{WS}& \grayref{[CVPR 2023]}&95HD&\rowcolor{shadecolor} 2.70& 2.79 & 6.68 & 8.24 & \\
&&ASD&\rowcolor{shadecolor} 0.79& 1.40 & 2.56 & 3.38 & \\
\cdashline{3-9}
% UniMatch + ENCORE&& 90.66& 88.73 & 87.82 & 82.07 & 77.69 & 85.74 & 84.97 & N/A \\
&& Dice& \textbf{90.07} & \textbf{88.68} & \textbf{85.81} & \textbf{84.88} & \textbf{29.05}\\
UniMatch + ENCORE && 95HD& 2.34 & 1.91 & 3.18  &3.55 & \\
&& ASD& 0.54 & 0.62 & 1.101 & 1.00 & \\

\specialrule{0.05em}{.05em}{.05em}

&& Dice &\rowcolor{shadecolor} \textbf{89.71} & \textbf{88.61} & 80.65  & 75.06  &25.20 \\
AD-MT \cite{AD-MT} & \grayref{[ECCV 2024]}& 95HD &\rowcolor{shadecolor} 3.15 & 2.62 & 6.22  & 20.36  & \\
&& ASD &\rowcolor{shadecolor} 0.93 & 0.83 & 1.67  & 7.87  & \\
\cdashline{3-9}
 && Dice& 89.54  & 88.21  & \textbf{85.61}  & \textbf{80.48}  & \textbf{27.65} \\
AD-MT + ENCORE && 95HD & 3.18  & 2.89  & 5.45   & 4.76   & \\
&& ASD & 0.93   & 0.84  &1.47  & 1.43   & \\
\specialrule{.12em}{.05em}{.05em} 
\end{tabular}%
}
\end{table}
% This is for Supplementary
\begin{table}[th!]
\centering
\caption{Comparisons with state-of-the-art methods on \textbf{LA} test set based on dice, 95HD, and ASD coefficients (\%).}
\label{tab:LA_sup}
\renewcommand{\arraystretch}{1.1}
\resizebox{1\columnwidth}{!}{%
\begin{tabular}{r*{1}@{\hspace{1.8pt}}l*{1}{>{\centering\arraybackslash}m{1.3cm}}*{5}{>{\centering\arraybackslash}m{1.4cm}}*{1}{>
{\centering\arraybackslash}m{0.5cm}}}
\specialrule{.12em}{.05em}{.05em}
Framework && Metric& 1/10 (8)& 1/20 (4) & 1/26 (3) & 1/40 (2) & Rel. Avg. \\
\specialrule{.12em}{.05em}{.05em}
&& Dice& 84.04& 60.86& 65.15& 33.79& N/A\\
supervised&& 95HD& 12.96& 43.75& 32.13& 59.82& \\
&& ASD & 2.28& 9.03& 3.59& 11.45&\\
\specialrule{0.05em}{.05em}{.05em}

&& Dice& \rowcolor{shadecolor} 89.52 & \textbf{87.53} & 85.67 & 84.24 & 25.78 \\
UniMatch \cite{WS}& \grayref{[CVPR 2023]}& 95HD& \rowcolor{shadecolor} 7.42 & 9.95 & 12.34 & 16.25 & \\
&& ASD& \rowcolor{shadecolor} 1.70 & 1.83 & 2.07 & 2.14 & \\
\cdashline{3-9}

&& Dice& \textbf{89.76}& 86.94 & \textbf{87.75} & \textbf{86.18} &\textbf{26.70} \\
UniMatch + ENCORE2 && 95HD&7.48 & 10.81 & 7.83 & 10.27 & \\
&& ASD &1.57 & 1.95 & 1.89 & 2.14 & \\
\specialrule{0.05em}{.05em}{.05em}
    
&& Dice& \rowcolor{shadecolor} \textbf{89.70}& 87.85& \textbf{88.84}& 86.60& 27.29 \\
Switch\cite{DualTeacher}&\grayref{[NeurIPS 2024]}& 95HD&\rowcolor{shadecolor} 7.12& 8.35& 7.44& 10.29&  \\
&& ASD&\rowcolor{shadecolor} 1.67& 1.91& 1.89& 2.50& \\

\cdashline{3-9}

&& Dice& 89.62&\textbf{88.66}&88.81&\textbf{86.73}& \textbf{27.50} \\
Switch + ENCORE2 &&95HD&7.35&7.40&7.18&9.78& \\
&&ASD&1.75&2.10&1.89&2.23& \\
\specialrule{0.05em}{.05em}{.05em}

& &Dice&\rowcolor{shadecolor} \textbf{90.01}  & \textbf{89.76} & 88.89& 83.21 & 27.01 \\
AD-MT \cite{AD-MT}& \grayref{[ECCV 2024]} &95HD&\rowcolor{shadecolor} 7.16  & 7.51 & 7.22& 11.37 &  \\
& &ASD&\rowcolor{shadecolor} 1.67  & 1.76 & 2.00& 2.79 & \\

\cdashline{3-9}

&&Dice&89.46 & 89.42 & \textbf{89.25} & \textbf{89.13} & \textbf{28.36} \\
AD-MT + ENCORE2 &&95HD&7.18 & 6.64 & 7.03 & 7.11 & \\
&&ASD&1.69 & 1.89 & 1.77 & 1.96 & \\

\specialrule{.12em}{.05em}{.05em}
\end{tabular}%
}
\end{table}

% \input{Tables/ablation_ADMT}
% \input{Tables/ablation_WeakStrong}

% \begin{figure}[!t]
%     \centering
%     \includegraphics[width=0.44\textwidth]{Figures/KDE_ACDC.png}
%     \caption{Kernel density estimates from various methods trained on 1 labeled ACDC dataset. From top to bottom, the plots display features corresponding to the right ventricle, myocardium, and left ventricle classes.}
%     \label{fig:kde}
% \end{figure}

% \input{Visualizations/figure}

Figure \ref{fig:thresholdvariation_weakstrong} illustrates how the Dice scores of the UniMatch model vary across different confidence thresholds on the ACDC dataset, under four labeled data fractions. As expected, using more labeled data leads to higher overall performance, with relatively small fluctuations in Dice values as the threshold changes. By contrast, when labeled data are extremely scarce (2\% or 1\%), the Dice scores are more sensitive to threshold selection, showing larger performance swings. This indicates that choosing an appropriate confidence threshold becomes increasingly crucial in low-labeled data settings, where pseudo-label quality must be carefully balanced to achieve optimal segmentation results.

\begin{figure*}[t!]
  \centering
  % First row (3 subfigures)
  \begin{subfigure}[t]{0.3\textwidth}
    \centering
    \includegraphics[width=1\textwidth]{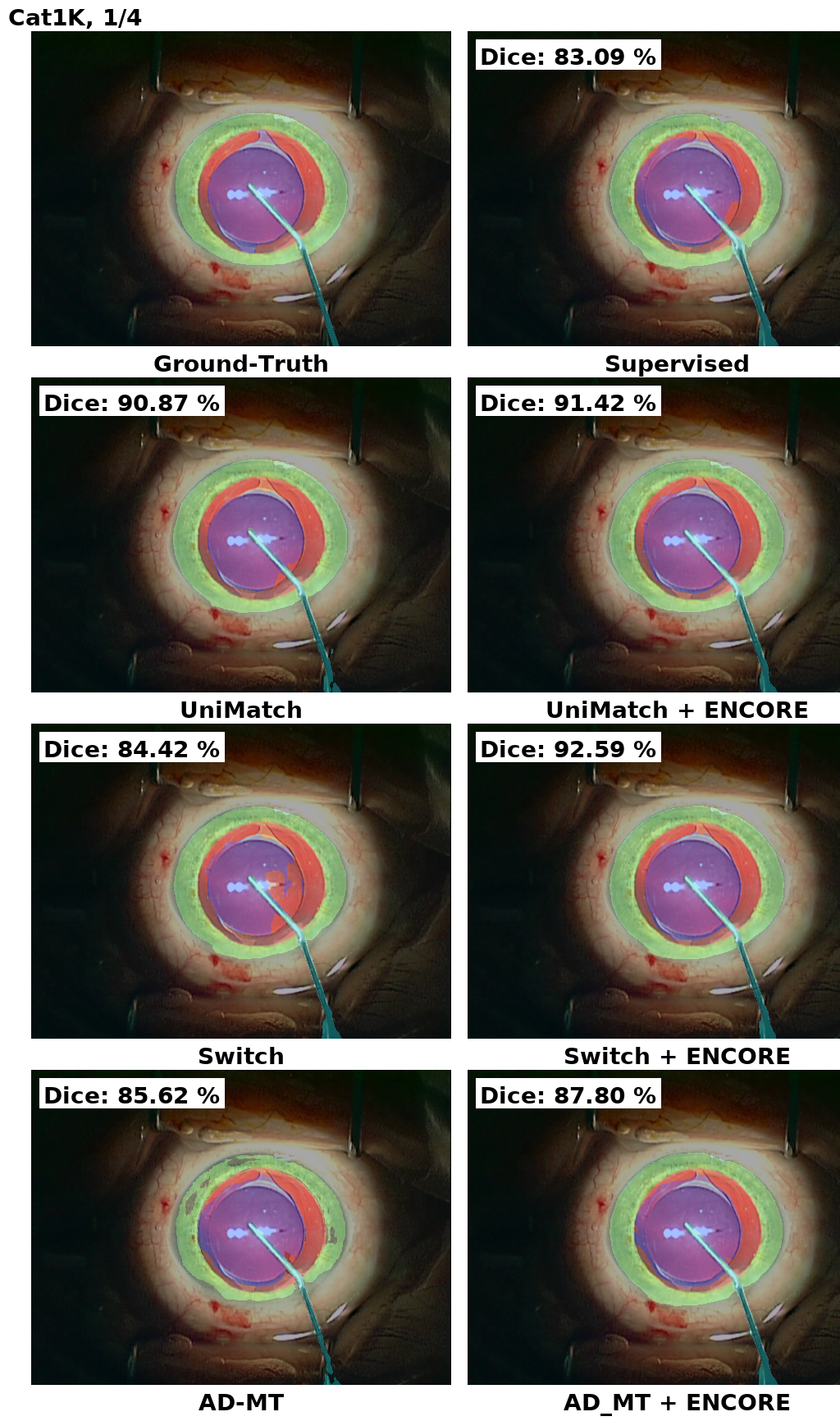}
  \end{subfigure}
  \begin{subfigure}[t]{0.3\textwidth}
    \centering
    \includegraphics[width=1\textwidth]{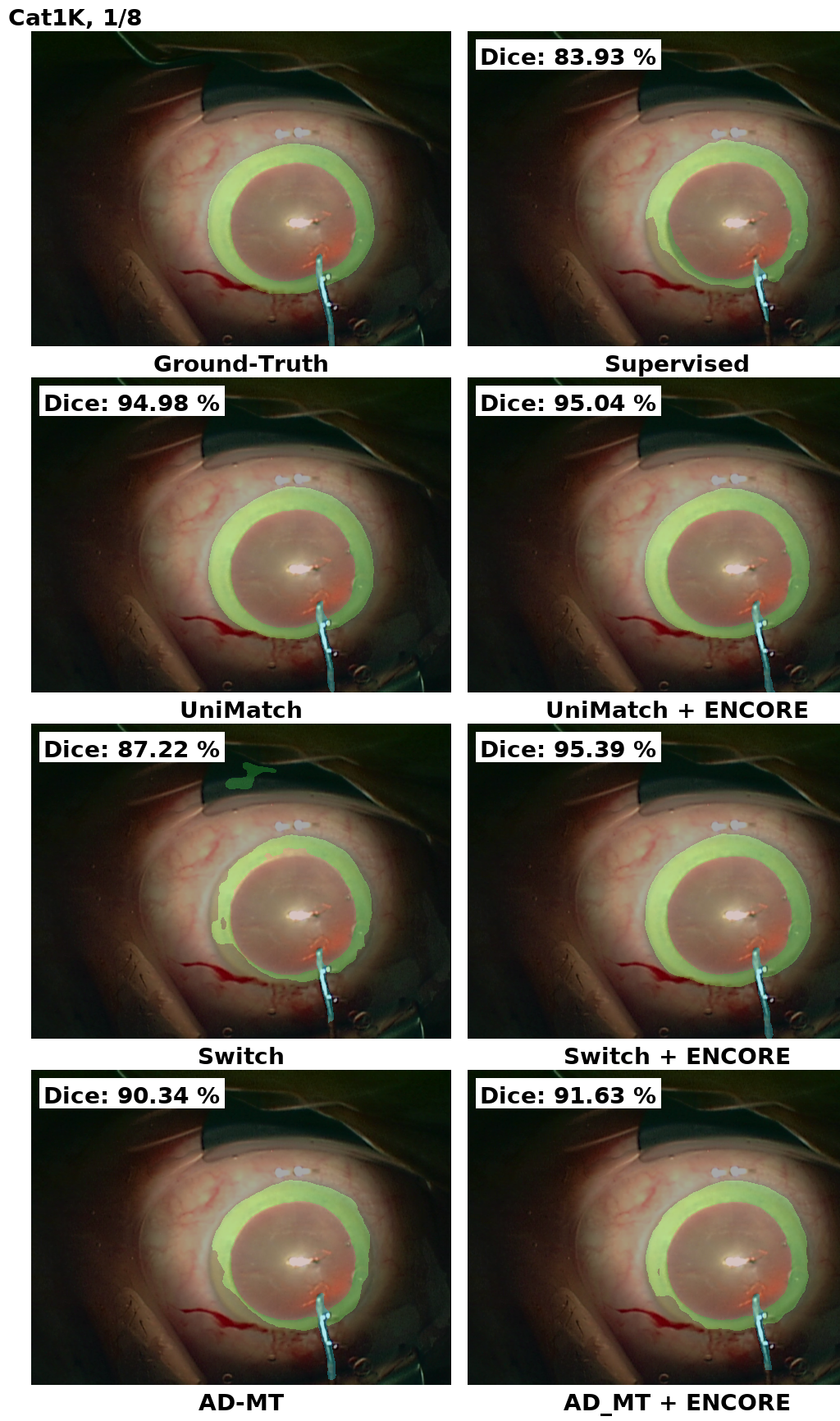}
  \end{subfigure}
  \begin{subfigure}[t]{0.3\textwidth}
    \centering
    \includegraphics[width=1\textwidth]{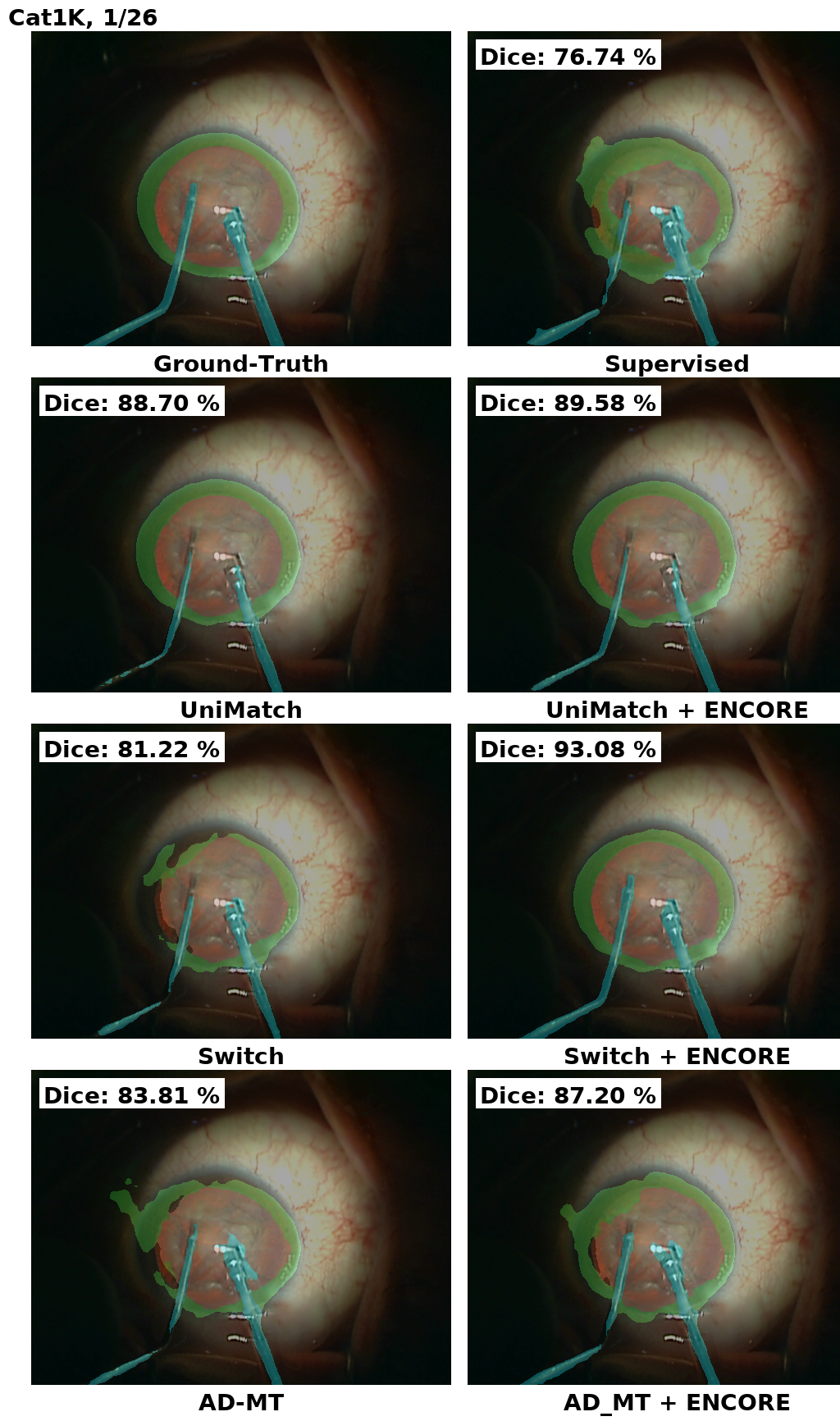}
  \end{subfigure}

  \caption{Qualitative comparisons of state-of-the-art methods with and without ENCORE on the Cataract-1K dataset across various labeled data fractions.}
  \label{fig:Cat1K_sup}
\end{figure*}

Tables \ref{tab:MRI_sup} and \ref{tab:EndoVis_sup} present extended results on the Prostate MRI and EndoVis datasets, respectively, for all tested data splits, complementing Tables 2 and 3 in the main paper. The proposed model consistently outperforms other semi-supervised approaches across every split. Notably, in Table \ref{tab:MRI_sup}, our UniMatch + ENCORE model achieves a Dice score of $66.82\%$ on 1/50 of the data, surpassing supervised training with 1/14 of the data ($65.05\%$). A similar pattern is observed in Table \ref{tab:EndoVis_sup} for the EndoVis dataset, which highlights the effectiveness of ENCORE in enhancing performance, particularly in low-label scenarios.

Tables \ref{tab:ACDC_sup} and \ref{tab:LA_sup} present extended evaluations for the ACDC and LA datasets, respectively, building on the results in the main paper. These tables include Dice, 95HD, and ASD metrics across various volume splits, illustrating that our ENCORE framework consistently surpasses baseline methods—particularly when labeled data are limited. Notably, in Table \ref{tab:ACDC_sup}, AD-MT+ENCORE achieves an 80.48\% Dice score using just one volume, outperforming the original AD-MT in Dice as well as 95HD and ASD. A similar trend emerges in the LA dataset (Table \ref{tab:LA_sup}), where ENCORE-based thresholding again demonstrates superior performance.

\begin{figure*}[t!]
  \centering
  % First row (3 subfigures)
  \begin{subfigure}[t]{0.3\textwidth}
    \centering
    \includegraphics[width=1\textwidth]{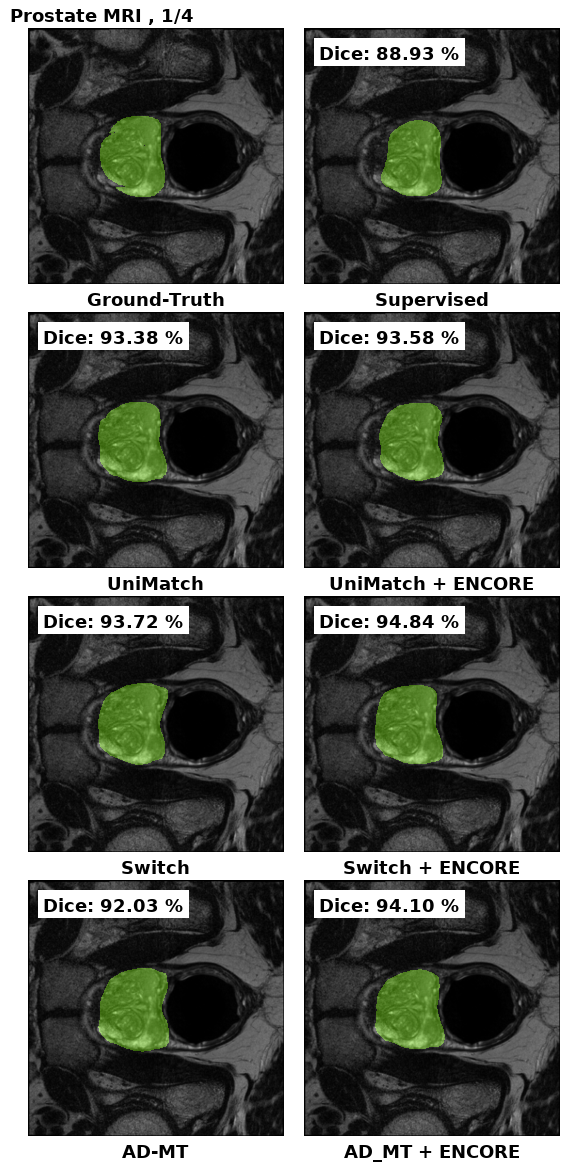}
  \end{subfigure}
  \begin{subfigure}[t]{0.3\textwidth}
    \centering
    \includegraphics[width=1\textwidth]{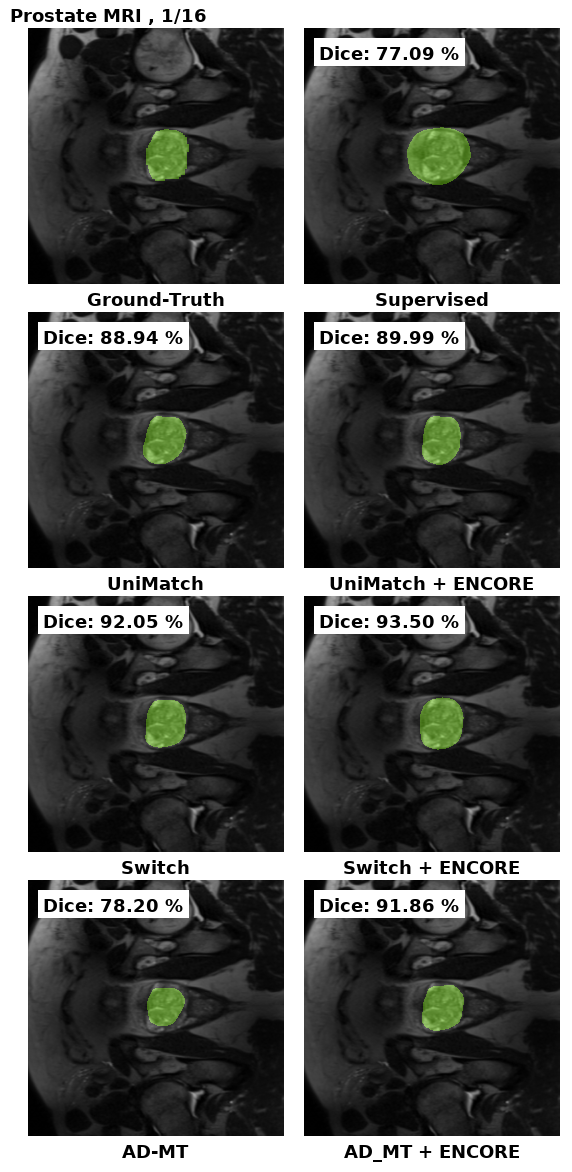}
  \end{subfigure}
  \begin{subfigure}[t]{0.3\textwidth}
    \centering
    \includegraphics[width=1\textwidth]{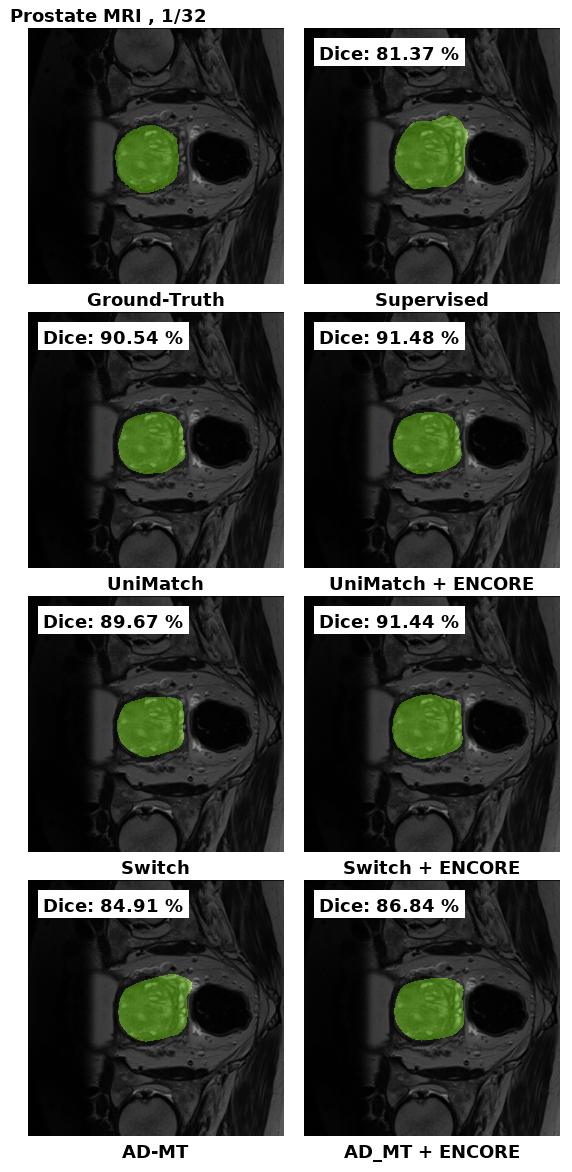}
  \end{subfigure}

  \caption{Qualitative comparisons of state-of-the-art methods with and without ENCORE on the Prostate MRI dataset across various labeled data fractions.}

  \label{fig:MRI_sup}
\end{figure*}

\begin{figure*}[t!]
  \centering
  % First row (3 subfigures)
  \begin{subfigure}[t]{0.3\textwidth}
    \centering
    \includegraphics[width=1\textwidth]{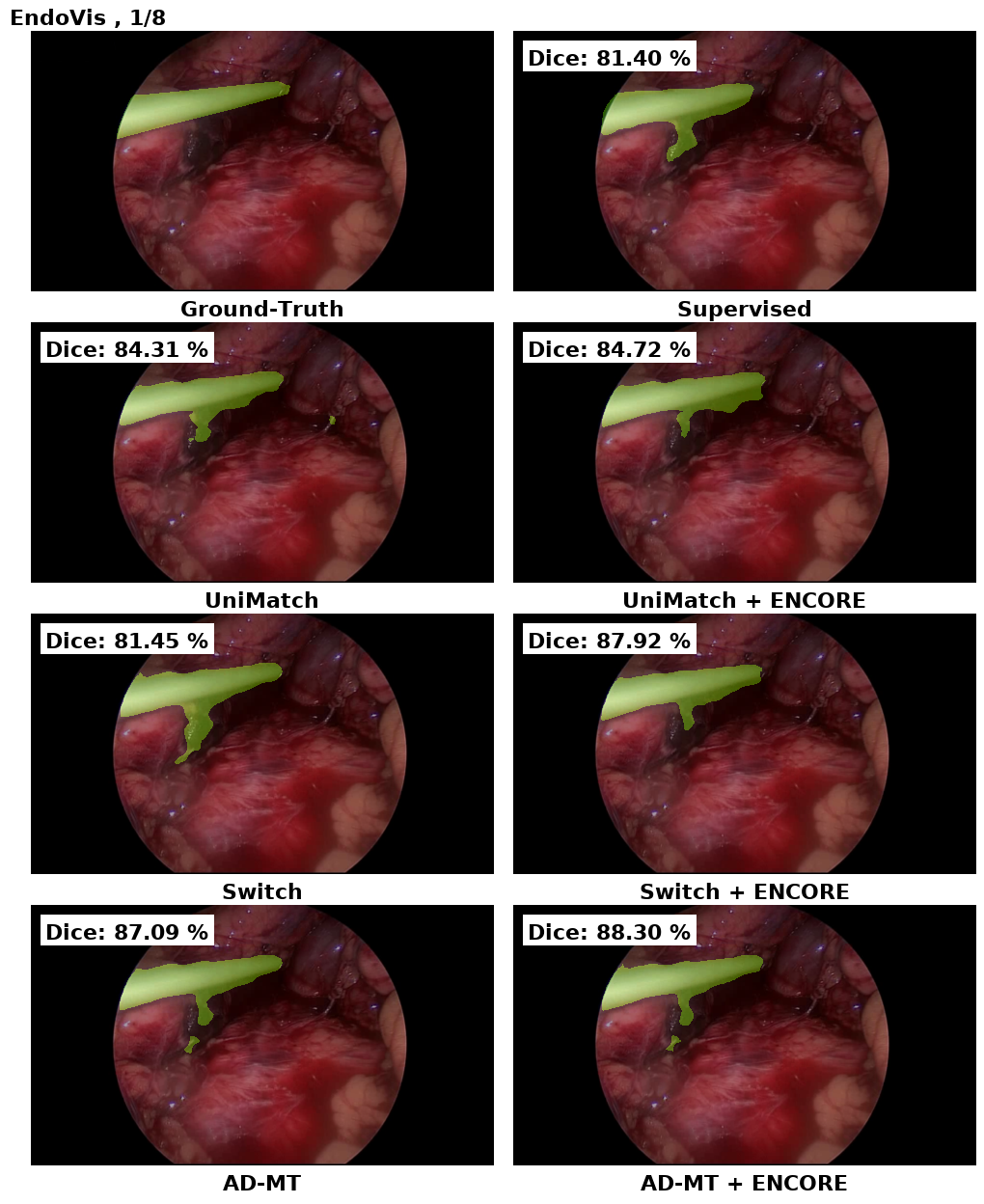}
  \end{subfigure}
  \begin{subfigure}[t]{0.3\textwidth}
    \centering
    \includegraphics[width=1\textwidth]{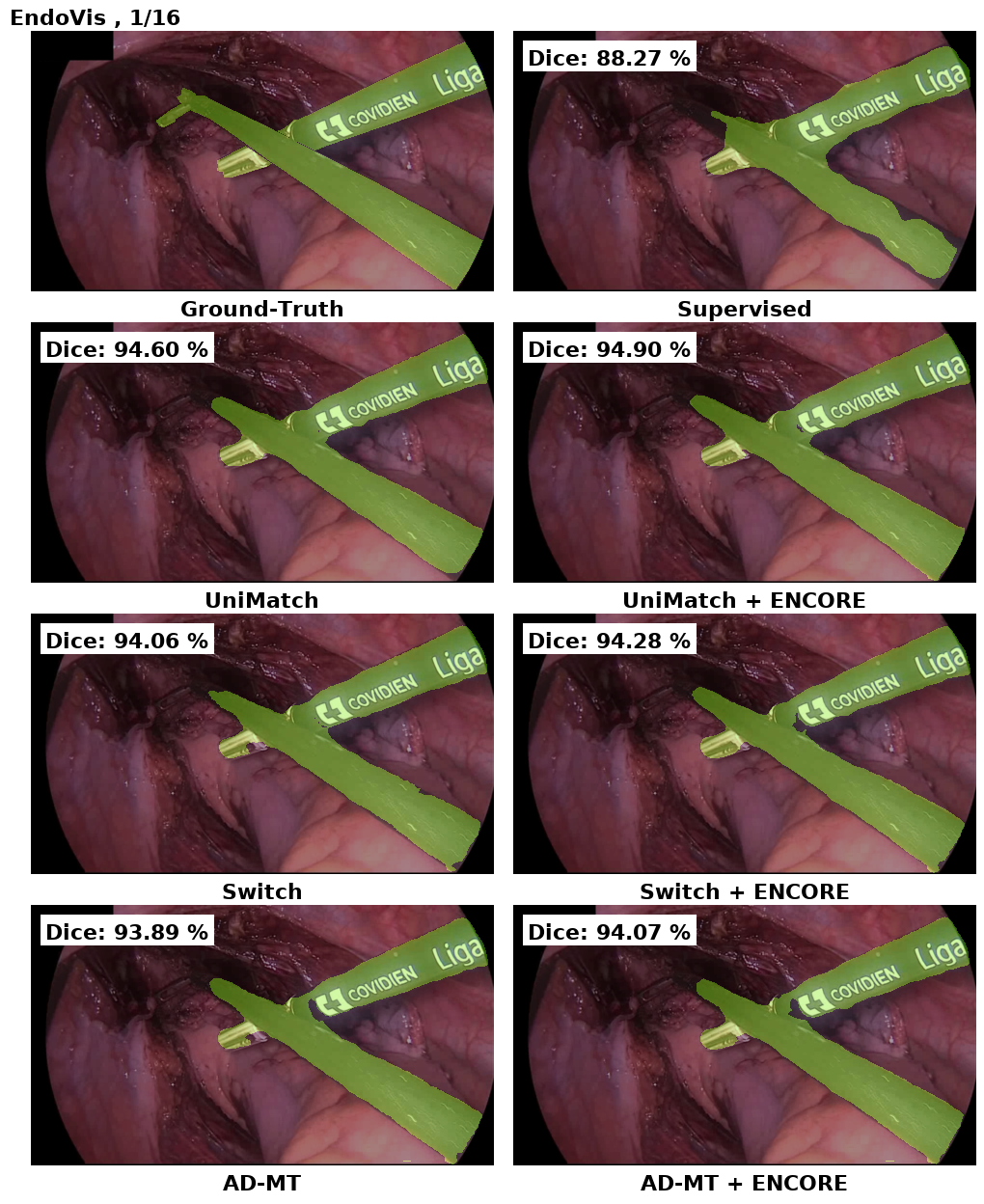}
  \end{subfigure}
  \begin{subfigure}[t]{0.3\textwidth}
    \centering
    \includegraphics[width=1\textwidth]{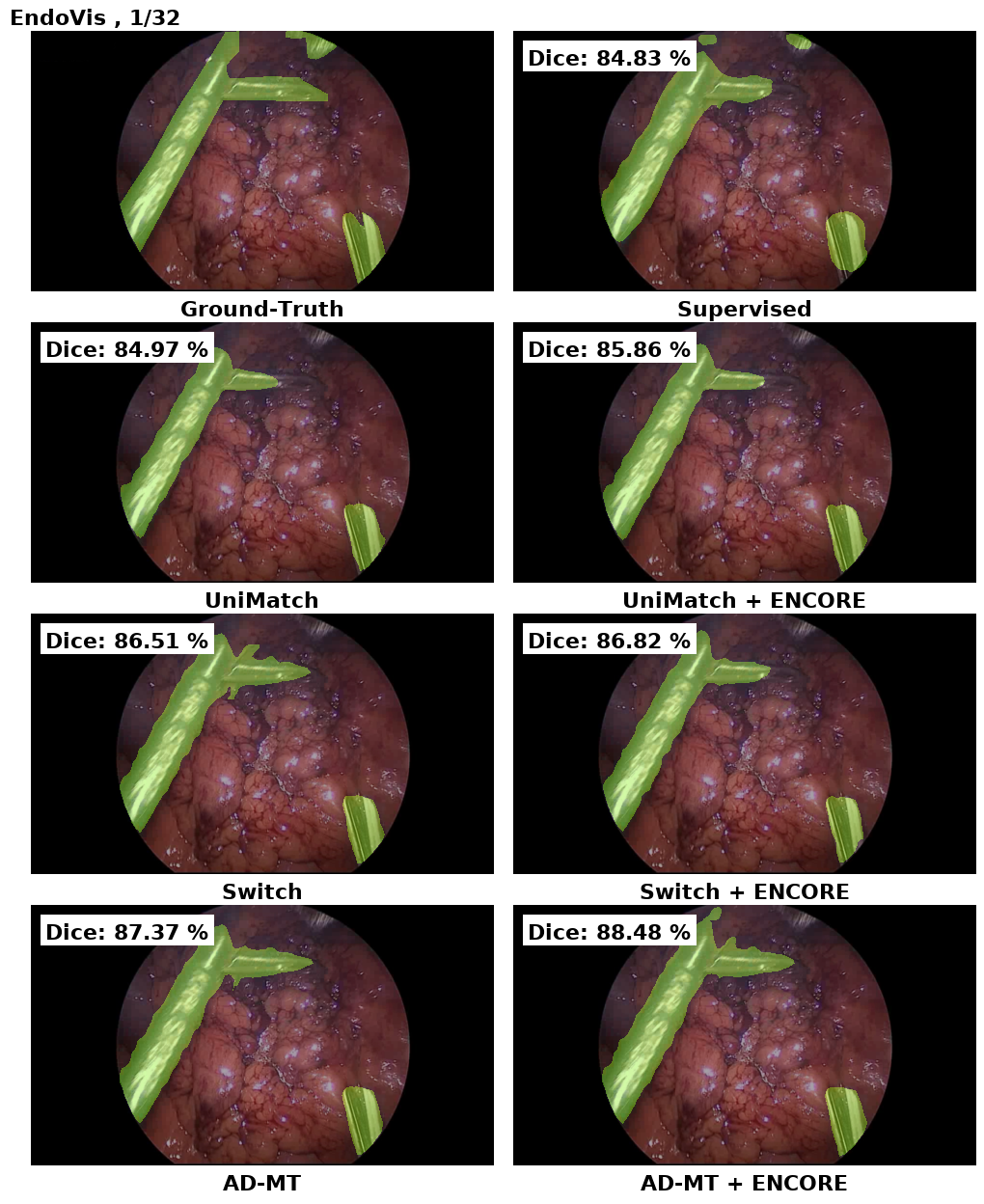}
  \end{subfigure}

  \caption{Qualitative comparisons of state-of-the-art methods with and without ENCORE on the EndoVis dataset across various labeled
data fractions.}
  \label{fig:Endovis_sup}
\end{figure*}

\begin{figure*}[!t]
    \centering
    \includegraphics[width=1\textwidth]{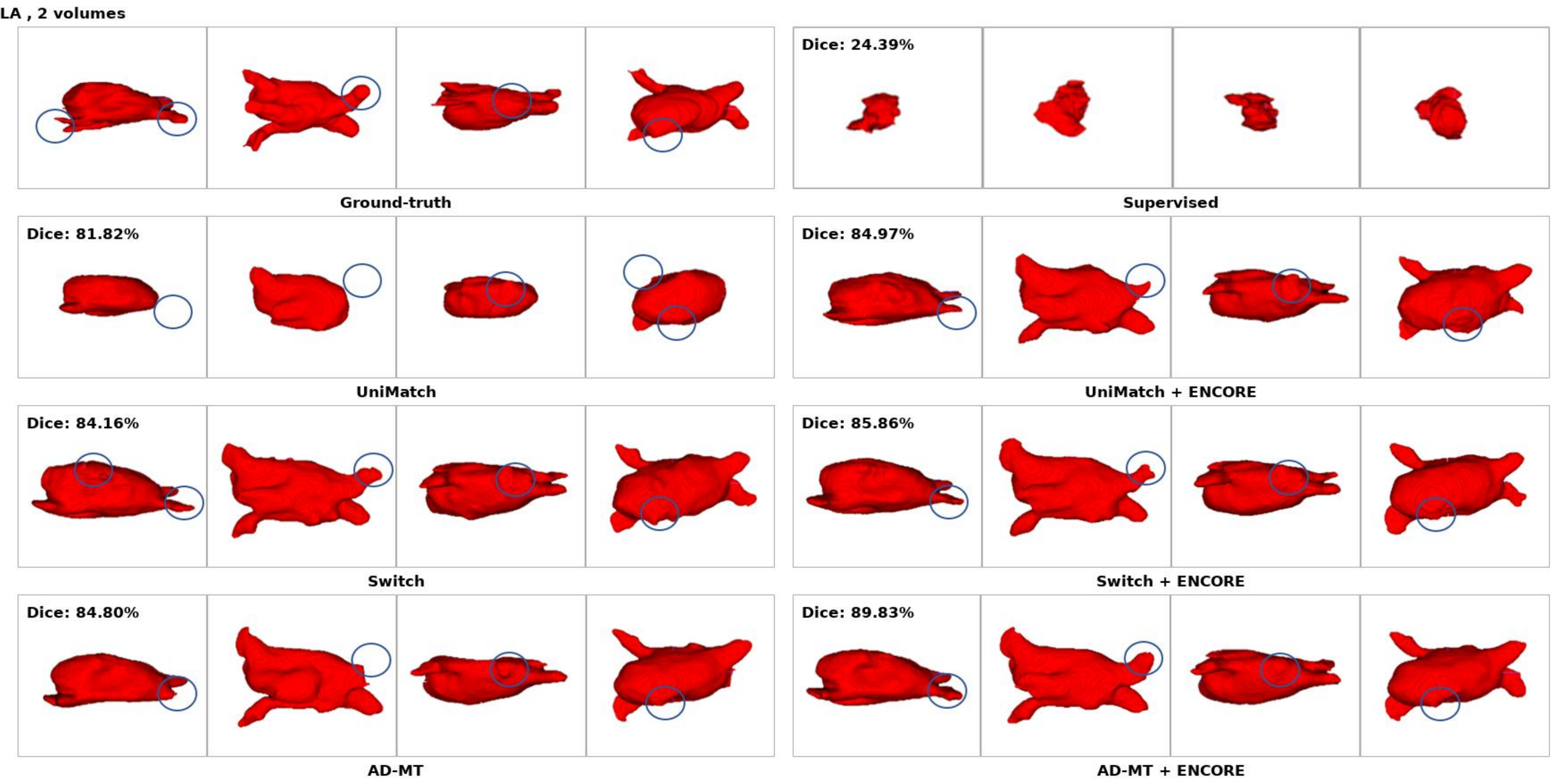}
    \caption{Qualitative comparisons of state-of-the-art methods with and without ENCORE on the LA dataset with 2 volume of labeled data.}
    \label{fig:LA_sup}
\end{figure*}

Figures \ref{fig:Cat1K_sup}, \ref{fig:MRI_sup}, \ref{fig:Endovis_sup}, and \ref{fig:LA_sup} show qualitative comparisons among the proposed approach, supervised learning, and three semi-supervised baselines (UniMatch \cite{WS}, Switch \cite{DualTeacher}, AD-MT \cite{AD-MT}). These evaluations cover different labeled data splits for four datasets: Cataract-1K, Prostate MRI, EndoVis, and LA. The visuals clearly indicate that all semi-supervised approaches significantly improve segmentation performance compared to supervised learning, especially when labeled data is limited. However, integration of our pseudo-label assessment method boosts the segmentation performance across all datasets. 
Figure \ref{fig:LA_sup} further illustrates four different views of the 3D LA dataset using models trained on only two volumes, where the ground truth comparisons highlight the superiority of ENCORE-based models over the baselines.

\begin{figure*}[!t]
    \centering
    \includegraphics[width=0.65\textwidth]{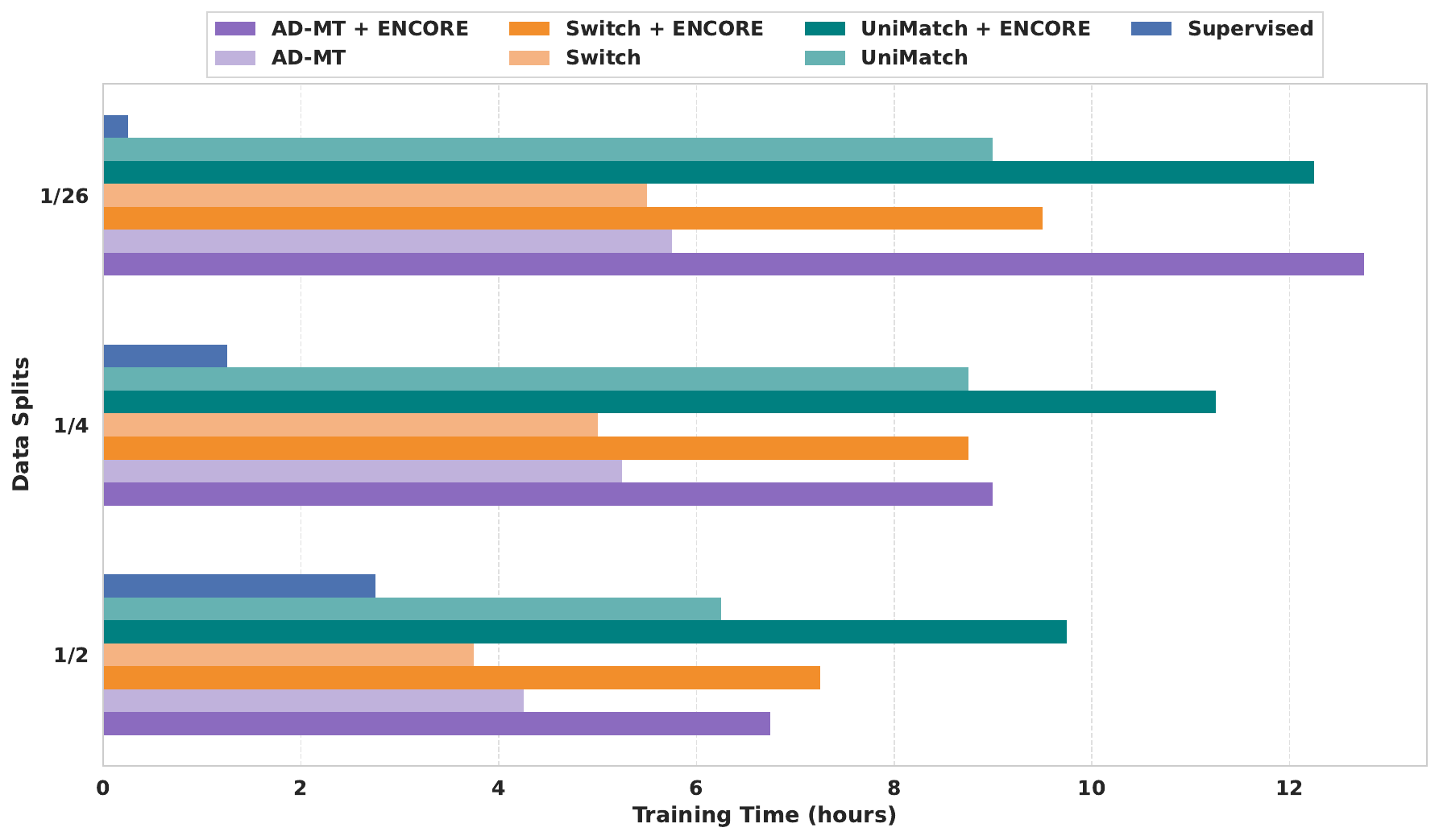}
    \caption{Training Time of Supervised and Semi-Supervised Methods on Catataract-1K dataset}
    \label{fig:training_time_sup}
\end{figure*}

% Figure \ref{fig:training_time_sup} compares the training times for our proposed models against other supervised and semi-supervised approaches on the Cat1K dataset, under different labeled data splits. Overall, the purely supervised model exhibits the shortest training time, as it does not leverage unlabeled data. In contrast, the semi-supervised methods—particularly the ENCORE models require longer training times due to additional iterative steps for refining unlabeled data.

Figure \ref{fig:training_time_sup} compares training times for our proposed feedback-driven strategy with supervised and semi-supervised approaches on the Cataract-1K dataset, across different labeled data splits. Overall, the purely supervised model exhibits the shortest training time, as it does not leverage unlabeled data. In contrast, the semi-supervised methods—especially those incorporating ENCORE—require additional iterative steps to assess pseudo labels, resulting in longer training durations. Although this increased computational cost can be considered a drawback, it remains comparable to other semi-supervised baselines, and the performance gains offered by ENCORE more than justify the overhead.

{
    \small
    \balance
    \bibliographystyle{ieeenat_fullname}
    \bibliography{main}
}